\documentclass{article}

\PassOptionsToPackage{table}{xcolor}
\usepackage[dblblindworkshop,final]{neurips_2026}

\usepackage[utf8]{inputenc} 
\usepackage[T1]{fontenc}    
\usepackage{hyperref}       
\usepackage{url}            
\usepackage{booktabs}       
\usepackage{amsfonts}       
\usepackage{nicefrac}       
\usepackage{microtype}      
\usepackage{microtype}
\usepackage{hyperref}
\usepackage{url}
\usepackage[most]{tcolorbox}
\usepackage{multirow}
\usepackage{graphicx}
\usepackage{booktabs}
\usepackage[table]{xcolor}
\usepackage{amsmath}
\usepackage{pifont}
\usepackage{algorithm}
\usepackage{algpseudocode}
\usepackage{float}
\usepackage{amsfonts}
\usepackage{tabularx}
\usepackage{multicol}
\usepackage{caption}
\usepackage{subcaption}
\usepackage{refcount}
\usepackage{makecell}
\usepackage{newfloat}
\usepackage{listings}
\usepackage{adjustbox}
\usepackage{wrapfig}
\usepackage{amssymb}
\usepackage{caption}
\usepackage{amsthm}
\usepackage{enumitem}
\theoremstyle{remark}
\usepackage{array}

\newtheorem{prop}{Proposition}
\newtheorem*{remark}{Remark}
\usepackage{bm}                       
\providecommand{\R}{\mathbb{R}}
\providecommand{\E}{\mathbb{E}}
\providecommand{\Sph}{\mathbb{S}^{d-1}}
\DeclareMathOperator*{\argmin}{arg\,min}
\DeclareMathOperator*{\argmax}{arg\,max}

\usepackage{lineno}

\definecolor{darkblue}{rgb}{0, 0, 0.5}
\hypersetup{colorlinks=true, citecolor=darkblue, linkcolor=darkblue, urlcolor=darkblue}

\workshoptitle{Linguistic Principles for Foundation Models (LP4FM)}
\title{Efficient Best-Of-$N$ with Radial Consensus Score}

\author{
Manh Nguyen\thanks{Corresponding Author}, \quad Sunil Gupta, \quad Hung Le \\
Applied Artificial Intelligence Initiative, Deakin University, Australia\\
\texttt{\{manh.nguyen, sunil.gupta, thai.le\}@deakin.edu.au}
}

\begin{document}

\maketitle

\begin{abstract}

Solving complex problems with large language models (LLMs) often requires selecting the best answer from multiple candidate responses (best-of-$N$), yet identifying the most reliable one remains challenging, particularly when correctness does not align with majority agreement.
Existing approaches based on voting or probability scoring rely on surface-level signals (e.g., frequency or token-level likelihood), often missing high-quality rare answers and ignoring the geometric structure of candidate representations, where correctness may emerge as semantic consistency rather than surface agreement.
To address these limitations, we introduce \textbf{Radial Consensus Score (RCS)}, a simple, efficient, and training-free method for best-of-$N$ selection. 
RCS models semantic consensus by computing a weighted Fr\'echet mean (semantic center) of answer embeddings and ranking candidates by their radial distance to this center. This enables it to recover semantically coherent correct answers despite lexical fragmentation, even when no single surface form holds a plurality under exact-match voting. 
Importantly, RCS offers a flexible framework that supports multiple weighting schemes, enabling the integration of arbitrary signals, such as semantic agreement or model confidence, with the position of the center adapting accordingly.
Extensive experiments across five open-weight models and various benchmarks demonstrate that RCS matches or outperforms strong baselines at every sampling budget, with the largest gains when correct answers are lexically fragmented.
Moreover, RCS serves as an effective drop-in replacement for majority voting in multi-agent debate and remains effective in black-box scenarios.
These results suggest that geometric consensus can serve as a scalable and reliable alternative to voting-based methods for answer selection.
Code is publicly available at \url{https://github.com/manhitv/RCS}.
\end{abstract}


\section{Introduction}

Large language models (LLMs) have achieved remarkable progress across diverse tasks, yet they remain prone to generating inconsistent or incorrect outputs, particularly in open-ended or reasoning-intensive scenarios~\citep{wang2022self, huang2025survey}. 
A common strategy to improve reliability is best-of-$N$ sampling: generate multiple candidate responses from the same prompt and select the highest-quality one according to a scoring mechanism~\citep{cobbe2021training, wang-etal-2024-math, kang2025scalable}. 
In this framework, performance critically depends not only on the diversity of sampled candidates but, more importantly, on the effectiveness of the selection criterion~\citep{maryanskyy2026agents}.

Existing selection methods can be broadly categorized into two groups. The first relies on \emph{external evaluators}, such as reward models or verifiers, to score each candidate~\citep{cobbe2021training, lightman2023let}. 
While effective, these approaches introduce additional computational overhead and require extra supervision. 
The second group uses \emph{intrinsic signals} derived from the sampled responses themselves. 
The dominant paradigm in this category is self-consistency via majority voting, which assumes that correct answers appear most frequently among samples~\citep{wang2022self}. 
This works well when the answer space is well-defined, but it struggles in settings with semantic ambiguity or when high-quality answers are rare despite being semantically close to the consensus~\citep{chen2023universal, kang2025scalable}.
Recent work has also proposed ranked or reasoning-aware extensions of self-consistency, such as Borda-style voting, along with theoretical analyses connecting Best-of-$N$ to mode estimation and KL-constrained optimization~\citep{gui2024bonbon, wang2025ranked, wan2025reasoning, cordero2025certified}. 
Parallel efforts explore weighted aggregation and hybrid selection strategies, such as correlation-aware multi-agent weighting and best-of-majority, to improve reliability and scaling behavior~\citep{ai2025beyond, di2025best}.

\begin{figure*}[t]
    \centering
    \includegraphics[width=\textwidth]{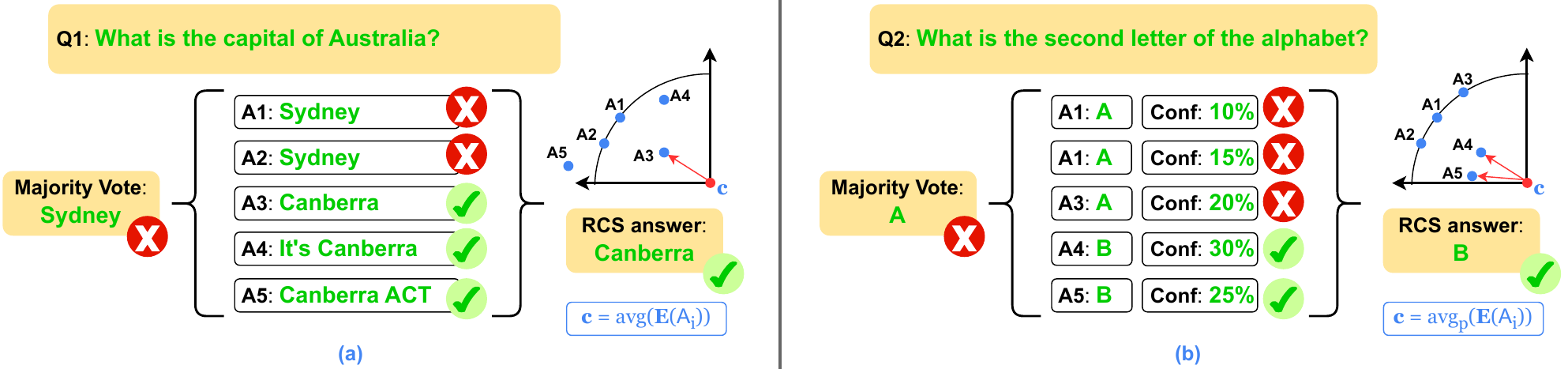}
    \caption{
        (a) Majority voting ignores semantically similar answers by treating surface forms independently.
        (b) It also fails to select high-confidence but minority answers.
        RCS addresses the lexical fragmentation issue by mapping answers into an embedding space via Encoder $\mathbf{E}$ and estimating a centroid $\mathbf{c}$, either \textit{uniformly} or \textit{probability-weighted} over answer embeddings; the latter can also recover high-confidence minority answers. Additional qualitative examples are provided in Section~\ref{sec:qualitative}.}
    \label{fig:overview}
\end{figure*}

However, these approaches often rely on additional sampling or surface-form agreement, limiting their practicality in real-world settings, especially when efficient, black-box-compatible, and training-free selection is required. 
As illustrated in Figure~\ref{fig:overview}, these limitations arise from a mismatch between surface-level signals and the underlying semantic structure of candidate answers.
This raises a central question: \emph{Can we design an efficient selection framework that operates directly in the answer space without external supervision or costly sampling?}

To tackle this question, we introduce the \textbf{Radial Consensus Score (RCS)}, a simple and effective geometric approach for efficient best-of-$N$ selection. 
Our key insight is to model semantic consensus explicitly via a weighted Fr\'echet mean~\citep{fletcher} as the semantic center of answer embeddings and then rank candidates by their radial distance to this center. 
This formulation naturally incorporates both semantic agreement structure and optional confidence signals (frequency, pairwise cosine similarity, or generation probability) through the choice of weighting distribution $P$.
Different choices of $P$ recover a spectrum of selection strategies, ranging from uniform weighting that treats all candidates equally to frequency-based, cosine-similarity-based, and probability-based schemes that emphasize consensus, mutual agreement, or model confidence. 
This flexibility allows the framework to adapt to different inference settings while remaining fully training-free and black-box compatible.
The resulting score is computationally lightweight, model-agnostic, and directly yields a ranking of individual candidates without requiring clustering or external verifiers.
We evaluate RCS on seven established benchmarks spanning short-form QA (SciQ, GPQA), mathematical reasoning (Arithmetics, GSM8K), and long-form multiple-choice tasks (MMLU Formal Logic, MMLU-Pro, BigBenchHard), using five diverse open-weight models. 
RCS matches or outperforms strong baselines at all tested sampling budgets, with gains are largest when correct answers are lexically fragmented and shrink on closed-form tasks, where RCS gracefully reduces to voting.
Additional analyses confirm their efficiency and robustness in black-box settings and multi-agent debate scenarios.
Our contributions are threefold:
\textbf{First}, we introduce Radial Consensus Score (RCS) as a geometric consensus method for best-of-$N$ selection that unifies probability- and semantic-based methods into a closed-form estimator without clustering, and provably outperforms voting in both infinite- and finite-sample regimes under mild assumptions.
\textbf{Second}, comprehensive experiments across diverse tasks and models demonstrate that RCS 
matches or outperforms strong baselines at every sampling budget and never trails self-consistency on average.
\textbf{Third}, ablation studies further confirm the effectiveness and applicability of RCS, showing its robustness in multi-agent debate and fully black-box settings, along with strong practical efficiency, underscoring its impact as a scalable aggregation method for LLM inference.

\section{Related Work}

\textit{Self-Consistency}, in its most common form, performs majority voting over $N$ sampled candidates~\citep{wang2022self}. 
Extensions such as universal and fine-grained self-consistency improve robustness but often incur extra LLM calls for evaluation or refinement~\citep{chen2023universal,wang2024integrate}. 
Alternatively, \textit{Self-Certainty} directly leverages the model's output probability distribution to estimate response quality without external reward models, showing strong scaling with $N$~\citep{kang2025scalable}.
Beyond voting-based strategies, recent advances include ranked and reasoning-aware aggregation, e.g. Borda-style voting~\citep{wang2025ranked, wan2025reasoning}. 
From a theoretical perspective, Best-of-$N$ has been analyzed under sampling and alignment objectives, linking it to mode estimation and KL-constrained optimization~\citep{gui2024bonbon, cordero2025certified}.


Geometric consensus itself is not new. Centroid-based MBR~\citep{deguchi2024centroid} and trajectory-level clustering~\citep{chen2024latent, choi2026modex} already exploit embedding geometry, and \textit{Semantic Consensus Weighting}~\citep{jain2023sc} computes pairwise cosine similarities and selects the highest-scoring trajectory group via a discrete aggregation step, thereby inheriting the minority-answer blindness of majority voting. These approaches differ mainly in \emph{what is embedded} and \emph{whether embeddings are grouped}. Under a linear (inner-product) utility, selecting the sample nearest the weighted centroid is exactly MBR decoding, making discrete grouping redundant. Our contribution is not the selection rule itself, but its application at the final-answer level through RCS, which recovers fragmented correct answers while avoiding the representation collapse and computational overhead of embedding full reasoning trajectories.

Uncertainty estimation methods offer a complementary lens: probability-based approaches exploit token likelihoods such as predictive entropy~\citep{guerreiro2022looking,manakul2023selfcheckgpt,nguyen2025probabilities}, while semantic entropy methods cluster responses in embedding space and compute entropy over cluster distributions~\citep{kuhn2023semantic,farquhar2024detecting,lin2023generating,nikitin2024kernel}. Both families, however, either ignore the model's native probability signals or are sensitive to clustering quality, and neither directly yields per-sample candidate scores for efficient ranking.

In contrast, RCS unifies these directions, bridging probabilistic and semantic perspectives via a closed-form weighted Fr\'echet mean at the \emph{final-answer} level, avoiding embedding-collapse or discrete grouping step while remaining efficient.

\section{Methodology}\label{sec:method}


\subsection{Problem Statement}\label{sec:problem}
Given a prompt $x$, we sample $N$ independent generations $\{y_1, \dots, y_N\}$ from a language model using multinomial decoding, where each $y_i$ denotes a full generated response. We then map each response to its final answer $\{a_i\}_{i=1}^N$, where $a_i$ is the extracted or normalized prediction implied by $y_i$, optionally associated with generation likelihoods $\{p(y_i \mid x)\}_{i=1}^N$.
Our goal is to identify the most reliable answer among the candidates. Formally, we consider the problem:
\begin{align}
    a^* = \arg\max_{a_i} \; Q(a_i|x),
\end{align}
where $Q(a_i|x)$ is an unknown latent quality score that ranks candidate answers according to their correctness or reliability for prompt $x$. Since $Q$ is not directly observable at inference time, it must be inferred from the sampled generations themselves.

\subsection{Answer Selection via Geometric Consensus}\label{sec:center}

To approximate the latent ranking signal $Q$, we model agreement among answers in a continuous semantic space. Formally, we encode each final answer into a vector space using a sentence embedding model $\mathbf{E}$, and then apply explicit $\ell_2$-normalization to obtain unit-norm vectors $\mathbf{u}_i \in \mathbb{R}^d$:
\begin{equation}
\mathbf{u}_i = \frac{\mathbf{E}(a_i)}{\|\mathbf{E}(a_i)\|_2}, 
\qquad \|\mathbf{u}_i\|_2 = 1.
\end{equation}
In the context of best-of-$N$, sampled answers form a distribution of semantic interpretations of the same query, and a representative solution should reflect their overall agreement. The Fr\'echet mean~\citep{fletcher} defines such a centroid as the point that minimizes the expected squared distance to all candidate embeddings in the semantic space. Given embeddings $\{\mathbf{u}_i\}_{i=1}^N \subset \mathbb{R}^d$ with a probability distribution $P{=}\{p_i\}_{i=1}^N$, the semantic center is defined as the Fr\'echet mean under squared Euclidean loss\footnote{see Appendix~\ref{app:proof_center} for details.}:
\begin{align}
    \mathbf{c}(P) = \arg\min_{\mathbf{z}} \sum_{i=1}^N p_i \|\mathbf{u}_i - \mathbf{z}\|_2^2,
        \label{eq:center}
\end{align}
This admits the closed-form solution:
\begin{align}
    \mathbf{c}(P) = \sum_{i=1}^N p_i \mathbf{u}_i.
\end{align}
We adopt the squared Euclidean loss for its simple closed-form solution. When the center is restricted to the discrete set $\{\mathbf{u}_i\}_{i=1}^N$, the problem reduces to selecting a representative sample:
\begin{align}
\mathbf{c}_{\text{medoid}}(P)
= \arg\min_{\mathbf{u}_j} \sum_{i=1}^N p_i \|\mathbf{u}_i - \mathbf{u}_j\|_2^2,
\label{eq:medoid}
\end{align}
which corresponds to a weighted medoid. This can be viewed as a discrete counterpart of the Fr\'echet mean, where the solution is constrained to be one of the observed samples.

Different choices of $P$ (Table~\ref{tab:rds_variants}) induce different notions of semantic consensus. In particular, the uniform distribution recovers the standard unweighted center, while frequency-based, similarity-based or likelihood-based distributions incorporate empirical or model-derived reliability signals.
Conceptually, $\mathbf{c}(P)$ represents a consensus embedding of the generated answers, where semantically similar answers pull the center toward their shared meaning and low-weight or noisy samples contribute less to the resulting center (see Figure~\ref{fig:overview}).

\subsection{Ranking via Radial Consensus Score}\label{sec:main-rcs}

Given a semantic center $\mathbf{c}(P)$, we define a ranking function based on the radial dispersion of each candidate embedding relative to this center.
For each candidate embedding $\mathbf{u}_i$, its \textbf{Radial Consensus Score} (RCS) is defined as:
\begin{align}
    \mathrm{RCS}_i(P) = \|\mathbf{u}_i - \mathbf{c}(P)\|_2.
\end{align}
RCS measures how semantically aligned each candidate is with the global consensus induced by the distribution $P$: smaller values indicate stronger agreement with the collective semantic structure, while larger values suggest divergence or potential noise. 
We adopt the $\ell_2$ norm for its simplicity and consistency with the center formulation (Eq.~\ref{eq:center}), as the choice of norm does not materially affect the relative ranking of candidates in practice.

\begin{table*}[t]
\centering
\caption{Different instantiations of the distribution $P$ in RCS, where $f_i$ and $\Pr(y_i|\mathbf{x})$ denote the empirical frequency and model probability of candidate $y_i$, respectively. For the cosine-similarity-weighted variant, $\cos(\mathbf{u}_i,\mathbf{u}_j)$ denotes the cosine similarity between $\ell_2$-normalized embeddings.}
\label{tab:rds_variants}
\begin{tabular}{l l l}
\toprule
\textbf{Variant} & \textbf{Distribution $P$} & \textbf{Supported LLMs} \\
\midrule
Uniform & $p_i = \frac{1}{N}$ & Black-box\\
Frequency-weighted & $p_i = \frac{f_i}{\sum_j f_j}$ & Black-box\\
Cosine-weighted & $p_i \propto \frac{1}{N}\sum_{j=1}^N \cos(\mathbf{u}_i,\mathbf{u}_j)$ & Black-box\\
Probability-weighted & $p_i \propto \Pr(y_i|\mathbf{x})$ & White-box\\
\bottomrule
\end{tabular}
\end{table*}

The final prediction is selected as the candidate with minimum radial dispersion to the estimated consensus:
\begin{align}
    a^*(P) = \arg\min_{a_i} \mathrm{RCS}_i(P).
\end{align}

When the consensus center is constrained to lie in the discrete candidate set, the final prediction is the candidate corresponding to the weighted medoid:
\begin{align}
    a^*(P) = a_{j^*}, \qquad \mathbf{u}_{j^*} = \mathbf{c}_{\text{medoid}}(P),
\end{align}
where $\mathbf{c}_{\text{medoid}}(P)$ is defined in Eq.~\ref{eq:medoid}.
This selects the candidate whose embedding minimizes the expected \emph{squared} distance to others under $P$, consistent with the squared-Euclidean objective (Eq.~\ref{eq:center}). When $P$ is uniform, this reduces to the standard medoid.

Overall, the choice of $P$ controls how consensus is formed: uniform weighting reflects pure geometric agreement, while frequency, cosine similarity, and probability weighting emphasizes stable, semantic agreement and high-confidence responses; all naturally suppressing inconsistent outputs. We refer to the corresponding variants as RCS$_\text{uni}$, RCS$_\text{freq}$, RCS$_\text{cosine}$, and RCS$_\text{prob}$. Notably, the first three depend only on model outputs and embeddings, making them readily applicable in black-box settings.
We provide the detailed pseudo-code for our method in Appendix Algorithm~\ref{alg:rds}.

\section{Theoretical Analysis}
\label{sec:theory}

We show that RCS recovers the correct answer whenever it forms the \emph{semantic}
majority, even when it is fragmented into several surface forms and therefore holds
no plurality under exact-match voting. All statements are on the unit sphere
$\Sph=\{\mathbf{x}\in\R^d:\|\mathbf{x}\|_2=1\}$, where $\ell_2$-normalized answer
embeddings live. Proofs are in Appendix~\ref{app:proof_prop1}
and~\ref{app:proof_prop2}.

\paragraph{Generative model.}
We model the $N$ candidate embeddings as i.i.d.\ draws from a distribution
$\mathcal D$ on $\Sph$ with a well defined mean direction,
\begin{align}
\label{eq:assumption}
\mathbf{u}_1,\dots,\mathbf{u}_N\overset{\mathrm{i.i.d.}}{\sim}\mathcal D
\ \text{on }\Sph,
\qquad
\E[\mathbf{u}]=\rho\,\mathbf{c}^*,\quad
\rho:=\|\E[\mathbf{u}]\|_2\in(0,1],\quad \mathbf{c}^*\in\Sph,
\end{align}
where $\mathbf{c}^*=\E[\mathbf{u}]/\|\E[\mathbf{u}]\|_2$ is the true semantic center
(a unit vector by construction) and $\rho$ measures concentration. We do
\emph{not} assume $\mathcal D$ is Gaussian or isotropic in $\R^d$: since
$\|\mathbf{u}\|_2=1$, a model $\mathbf{u}=\mathbf{c}^*+\boldsymbol\varepsilon$ would
force $\|\E[\mathbf{u}]\|=\E\|\mathbf{u}\|=1$, which by Jensen's inequality holds
only for $\boldsymbol\varepsilon\equiv\mathbf 0$; Assumption~\eqref{eq:assumption}
avoids this by placing $\E[\mathbf{u}]$ strictly inside the sphere. It holds for the
von Mises Fisher law, for any distribution rotationally symmetric about
$\mathbf{c}^*$, and for anisotropic $\mathcal D$ with nonzero resultant. The
weighting $P=\{p_i\}$ (with $\sum_i p_i=1$) defines the center
$\mathbf{c}(P)=\sum_i p_i\mathbf{u}_i$ and covers all four variants of
Table~\ref{tab:rds_variants}. Because $\|\mathbf{u}_i\|_2=1$, RCS's rule is purely
directional,
\begin{align}
\label{eq:selection-identity}
\argmin_i\|\mathbf{u}_i-\mathbf{c}(P)\|_2=\argmax_i\langle\mathbf{u}_i,\mathbf{c}(P)\rangle,
\end{align}
so only the direction of $\mathbf{c}(P)$ matters and the analysis is unaffected by
the shrinkage $\rho$.

\begin{prop}[\textbf{Asymptotic recovery}]
\label{prop:asymptotic}
Under Assumption~\eqref{eq:assumption}, for all four weighting variants the center is
directionally consistent, $\mathbf{c}(P)/\|\mathbf{c}(P)\|_2\xrightarrow{a.s.}\mathbf{c}^*$
(for the cosine and probability variants, provided $\mathcal D$ is rotationally
symmetric about $\mathbf{c}^*$ and, for probability, that the model likelihood
increases with alignment to $\mathbf{c}^*$). Consequently
$\hat i\xrightarrow{a.s.}\argmax_i\langle\mathbf{u}_i,\mathbf{c}^*\rangle$ and
$\Pr(\text{RCS selects the correct answer})\to1$ as $N\to\infty$, whereas
exact-match voting selects the correct semantic cluster with probability at most
$1/L$ when that cluster is split across $L\ge2$ surface forms with no plurality.
Thus RCS is asymptotically superior whenever surface fragmentation occurs.
\end{prop}
\begin{proof}
See Appendix~\ref{app:proof_prop1}.
\end{proof}

\begin{prop}[\textbf{Finite sample gap}]
\label{prop:finite}
Let the candidates split into cluster $A$ (correct, size $k$, mean direction
$\boldsymbol\mu_A$) and cluster $B$ (spurious, size $N-k$, mean direction
$\boldsymbol\mu_B$), with common concentration $\rho$ and angular separation
$\theta=\angle(\boldsymbol\mu_A,\boldsymbol\mu_B)$. Define the signed margin
$\Delta=\rho\sin(\theta/2)\,(2k-N)/N$. For the geometric group (uniform and
frequency), if $\Delta>0$ then
\begin{align}
\label{eq:finite-bound}
\Pr(\text{RCS selects }A)\ \ge\ 1-\exp\!\big(-N\Delta^2/2\big).
\end{align}
Uniform weighting gives $\Delta>0\iff k>N/2$; frequency weighting replaces
$(2k-N)/N$ by the signed weighted mass, recovering a lexically dominant correct
cluster even at $k<N/2$. Cosine weighting has no finite sample guarantee, as it
amplifies the tightest cluster regardless of correctness; probability weighting can
recover a genuine semantic minority when its likelihood aligns with correctness.
When $\Delta<0$, RCS follows the majority and fails as voting does, while asymptotic
recovery (Proposition~\ref{prop:asymptotic}) still holds.
\end{prop}
\begin{proof}
See Appendix~\ref{app:proof_prop2}.
\end{proof}

\section{Experiment}\label{sec:experiment}

\begin{table}[t]
  \centering
  \caption{Best-of-$N$ selection accuracy. \textbf{BBF}: black-box friendly. For non-RCS methods, \underline{underline} marks the best baseline per setting. Among RCS variants, \textbf{bold} indicates the best RCS method per column. 
  Item-level McNemar tests (Appendix Table~\ref{tab:r9-full-per-cell}) show that RCS$_\text{uni}$ significantly outperforms self-consistency in 15 of 25 model$\times$benchmark cells ($p<0.05$).
  Additional results in Appendix Table~\ref{tab:ranking-addition}.}
  \label{tab:ranking}
  \resizebox{\columnwidth}{!}{
  \begin{tabular}{ll |c| ccccc | c}
    \toprule
    \textbf{Model} & \textbf{Method} & \textbf{BBF}
    & \textbf{SciQ} & \textbf{GPQA} 
    & \textbf{Arithmetics} & \textbf{GSM8K} & \textbf{Form.Log.} 
    & \textbf{Average}
    \\
    \midrule
    \multirow{13}{*}{Qwen2.5-3B}
    & Greedy & \ding{51} & 59.3 & 20.9 & 89.0 & 65.0 & 34.9 & 53.8\\
    & Oracle & \ding{51} & 80.8$\pm$0.7	& 51.0$\pm$2.9 & 99.5$\pm$0.5	& 93.1$\pm$0.8 &	73.5$\pm$5.7 & 79.6\\
    \cmidrule(lr){2-9}
    & NLL & \ding{55}& 59.4$\pm$0.7 & 15.7$\pm$2.4 & 36.0$\pm$0.9 & 25.0$\pm$3.1 & 23.0$\pm$2.7 & 31.8 \\
    & ANLL & \ding{55}& 59.2$\pm$0.5 & 15.9$\pm$2.2 & 36.0$\pm$1.3 & 24.9$\pm$3.3 & 22.8$\pm$3.0 & 31.8 \\
    & SC & \ding{51}   & 64.4$\pm$0.6 & 21.4$\pm$0.6 & 78.7$\pm$3.3 & 52.8$\pm$1.3 & 40.5$\pm$2.1 & 51.6 \\
    & SCW & \ding{51} & \underline{64.5$\pm$0.8} & \underline{23.0$\pm$0.3} & 51.8$\pm$3.0&	37.4$\pm$2.8	&25.7$\pm$2.4 & 40.5 \\
    & CE & \ding{55}   & 64.0$\pm$0.5 & 16.8$\pm$1.7 & \underline{80.3$\pm$4.3} & \underline{53.3$\pm$2.8} & \underline{41.3$\pm$0.8} & 51.1 \\
    & ModeX & \ding{51} & 62.3$\pm$0.1	&18.1$\pm$2.3	&48.5$\pm$2.3&	33.3$\pm$3.1	&25.1$\pm$2.6 & 37.5\\
    &\cellcolor{green!15} RCS$_\text{medoid}$ &\cellcolor{green!15} \ding{51} &\cellcolor{green!15} 65.2$\pm$1.0 &\cellcolor{green!15} 23.8$\pm$0.5 &\cellcolor{green!15} 81.7$\pm$3.0 &\cellcolor{green!15} 63.6$\pm$2.6 &\cellcolor{green!15} 43.6$\pm$1.6 &\cellcolor{green!15} 55.6 \\
    &\cellcolor{green!15} RCS$_\text{uni}$ &\cellcolor{green!15} \ding{51} & \cellcolor{green!15} \textbf{65.6$\pm$0.4} & \cellcolor{green!15} 23.1$\pm$2.2 & \cellcolor{green!15} \textbf{85.7$\pm$3.2} &\cellcolor{green!15} \textbf{65.8$\pm$3.1} &\cellcolor{green!15} \textbf{44.7$\pm$1.2} &\cellcolor{green!15} \textbf{57.0} \\
    & \cellcolor{green!15} RCS$_\text{freq}$ &\cellcolor{green!15} \ding{51} &\cellcolor{green!15} 64.8$\pm$0.7 &\cellcolor{green!15} 22.8$\pm$0.5 &\cellcolor{green!15} 82.8$\pm$4.6 &\cellcolor{green!15} 59.7$\pm$1.6 &\cellcolor{green!15} 43.4$\pm$1.7 &\cellcolor{green!15} 54.7 \\
    & \cellcolor{green!15} RCS$_\text{cosine}$ &\cellcolor{green!15} \ding{51} &\cellcolor{green!15} \textbf{65.6$\pm$0.6} &\cellcolor{green!15} \textbf{24.2$\pm$1.8} &\cellcolor{green!15} 85.2$\pm$4.5 &\cellcolor{green!15} 65.0$\pm$2.8 &\cellcolor{green!15} \textbf{44.7$\pm$1.2} &\cellcolor{green!15} 56.9 \\
    & \cellcolor{green!15} RCS$_\text{prob}$ &\cellcolor{green!15} \ding{55} &\cellcolor{green!15} 65.5$\pm$0.6 &\cellcolor{green!15} 22.5$\pm$2.4 &\cellcolor{green!15} 84.2$\pm$4.0&\cellcolor{green!15} 63.4$\pm$1.8 &\cellcolor{green!15} 43.9$\pm$1.2 &\cellcolor{green!15} 55.9 \\
    \midrule

    \multirow{13}{*}{Llama3.2-3B}
    & Greedy & \ding{51}& 55.6 & 23.0 & 96.0 & 79.7 & 34.9 & 58.3 \\
    & Oracle & \ding{51} & 79.8$\pm$0.4 &	52.5$\pm$1.5 & 99.8$\pm$0.3	& 92.6$\pm$0.3	& 71.4$\pm$4.4 & 79.2\\
    \cmidrule(lr){2-9}
    & NLL & \ding{55}& 53.6$\pm$1.1 & 20.2$\pm$3.2 & 86.5$\pm$1.3 & 74.3$\pm$1.5 & \underline{33.3$\pm$2.1} & 53.6 \\
    & ANLL & \ding{55}& 53.6$\pm$0.5 & 19.7$\pm$0.5 & 88.5$\pm$2.6 & 73.9$\pm$2.4 & 29.9$\pm$6.0 & 53.1 \\
    & SC & \ding{51}   & 58.8$\pm$1.1 & 17.4$\pm$2.6 & \underline{97.5$\pm$1.3} & 78.0$\pm$0.8 & 16.1$\pm$3.2 & 53.6 \\
    & SCW & \ding{51} & \underline{59.5$\pm$1.0} & \underline{25.4$\pm$0.9} & 69.2$\pm$2.0	&62.6$\pm$1.1	&18.5$\pm$3.3 & 47.0 \\
    & CE & \ding{55}   & 59.3$\pm$0.6 & 20.9$\pm$2.2 & 97.3$\pm$1.3 & \underline{78.5$\pm$0.8} & 17.5$\pm$2.7 & 54.7 \\
    & ModeX & \ding{51} & 53.2$\pm$0.6&	18.1$\pm$2.1	&70.0$\pm$2.3	&62.6$\pm$2.1	&16.9$\pm$0.5& 44.2\\
    &\cellcolor{green!15} RCS$_\text{medoid}$ &\cellcolor{green!15} \ding{51} &\cellcolor{green!15} \textbf{60.0$\pm$1.2} &\cellcolor{green!15} 25.6$\pm$0.8 &\cellcolor{green!15} \textbf{98.3$\pm$1.0} &\cellcolor{green!15} 80.5$\pm$1.5 &\cellcolor{green!15} 19.6$\pm$3.7 &\cellcolor{green!15} 56.8 \\
    &\cellcolor{green!15} RCS$_\text{uni}$ &\cellcolor{green!15} \ding{51} &\cellcolor{green!15} 59.5$\pm$0.8 &\cellcolor{green!15} 25.0$\pm$0.8 &\cellcolor{green!15} 98.2$\pm$1.0 &\cellcolor{green!15} 80.5$\pm$1.5 &\cellcolor{green!15} 20.4$\pm$3.2 &\cellcolor{green!15} 56.7 \\
    & \cellcolor{green!15} RCS$_\text{freq}$ &\cellcolor{green!15} \ding{51} &\cellcolor{green!15} 59.8$\pm$1.0 &\cellcolor{green!15} 25.4$\pm$0.9 &\cellcolor{green!15} 98.2$\pm$1.0 &\cellcolor{green!15} 79.4$\pm$1.3 &\cellcolor{green!15} 18.0$\pm$3.7 &\cellcolor{green!15} 56.2 \\
    & \cellcolor{green!15} RCS$_\text{cosine}$ &\cellcolor{green!15} \ding{51} &\cellcolor{green!15} 59.8$\pm$1.2 &\cellcolor{green!15} 25.2$\pm$1.2 &\cellcolor{green!15} 98.0$\pm$0.9 &\cellcolor{green!15} 81.1$\pm$1.6 &\cellcolor{green!15} 20.6$\pm$3.6 &\cellcolor{green!15} 56.9 \\
    & \cellcolor{green!15} RCS$_\text{prob}$ &\cellcolor{green!15} \ding{55} &\cellcolor{green!15} 59.6$\pm$0.8 &\cellcolor{green!15} \textbf{26.2$\pm$1.3} &\cellcolor{green!15} 97.5$\pm$0.8 &\cellcolor{green!15} \textbf{84.1$\pm$0.8} &\cellcolor{green!15} \textbf{33.1$\pm$3.2} &\cellcolor{green!15} \textbf{60.2} \\

    \bottomrule
  \end{tabular}
  }
\end{table}

\subsection{Experiment Setup}\label{sec:experiment-setup}

\paragraph{Models And Datasets.} We evaluate on five popular open-weight models from distinct families: Qwen2.5-3B, 7B~\citep{yang2025qwen25}, Llama3.2-3B, Llama3.1-8B~\citep{grattafiori2024llama}, and Gemma2-9B~\citep{team2024gemma}. We use seven established benchmarks covering both short-form question-answering and long-form reasoning tasks with different answer formats: (1) Short-form QA: \textbf{SciQ}~\citep{welbl2017crowdsourcing} and GPQA Diamond (\textbf{GPQA}~\citep{rein2024gpqa}), (2) Long-form Math Reasoning: \textbf{Arithmetics}~\citep{choi2025debate}, \textbf{GSM8K} \citep{cobbe2021training}, and (3) Long-form QA multiple choice: MMLU Formal Logic (\textbf{Form.Log.}~\citep{hendrycks2020aligning}), \textbf{MMLU-Pro}~\citep{wang2024mmlu}, BigBenchHard (\textbf{BBH-Date}, \textbf{BBH-Navigate}~\citep{suzgun2023challenging}).

\paragraph{Baselines.} 

We compare our method against the following baselines, which require only single generation pass:
(1) \textbf{NLL}~\citep{nguyen2025probabilities}, which selects answers with the highest likelihood based on negative log-likelihood;
(2) \textbf{ANLL}~\citep{guerreiro2022looking}, which uses average negative log-likelihood for length-normalized scoring;
(3) Self-Consistency (\textbf{SC}~\citep{wang2022self}), selecting the most frequent answer; 
(4) Semantic Consensus Weighting (\textbf{SCW}~\citep{jain2023sc}), which embeds full reasoning paths, computes pairwise cosine similarities, groups scores by unique reasoning path, and selects the highest-scoring group; 
(5) Self-Certainty (\textbf{CE}~\citep{kang2025scalable}), which ranks answers by combining frequency and model confidence using Borda voting; 
and (6) \textbf{ModeX}~\citep{choi2026modex}, which applies spectral clustering over reasoning trajectories and selects answers from the resulting cluster centroids. 
We report four RCS variants (Table~\ref{tab:rds_variants}): \textbf{RCS$_\text{uni}$}, \textbf{RCS$_\text{freq}$}, \textbf{RCS$_\text{cosine}$}, and \textbf{RCS$_\text{prob}$} (using ANLL as a length-normalized probability proxy). We also report \textbf{RCS$_\text{medoid}$} (uniform, unweighted) as a discrete baseline. 
\begin{figure*}[ht]
    \centering
    \includegraphics[width=\textwidth]{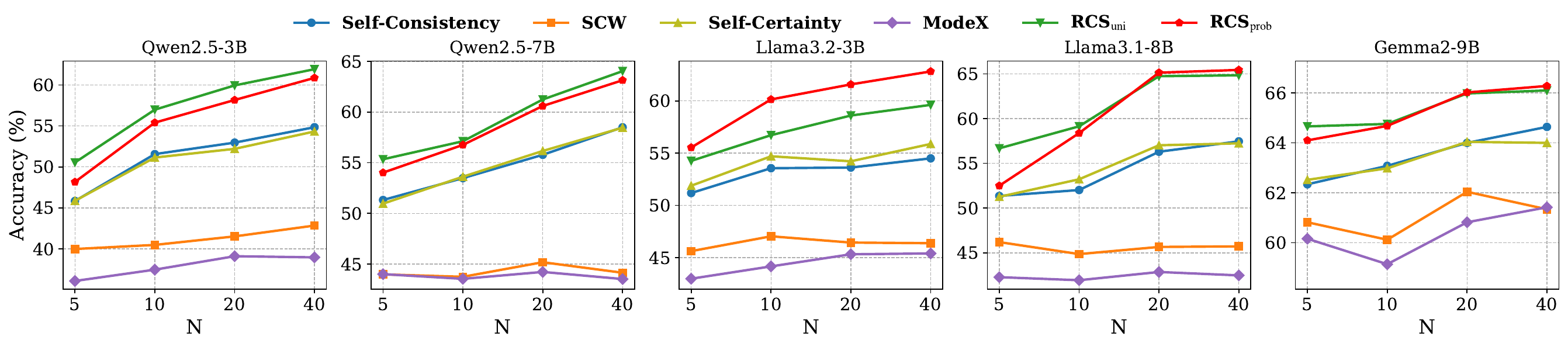}
    \caption{
    Average performance over five benchmarks for different numbers of sampling responses $N$. Each subplot corresponds to an LLM backbone, showing how accuracy changes with increasing $N$.}
    \label{fig:scaling}
\end{figure*}

\begin{table}[ht]
  \centering
  \caption{Performance on Arithmetics and Form.Log. with Multi-agent Debate. Best values are bolded. \textbf{R=0,1,2} indicate debate rounds, while \textbf{Avg.} denotes the average over all rounds.}
  \label{tab:mad_qwen2.5-3b}
  \resizebox{\columnwidth}{!}{
  \begin{tabular}{cll|cccc|cccc}
    \toprule
    \multirow{2}{*}{\textbf{Model}} 
    & \multirow{2}{*}{\textbf{Method}} 
    & \multirow{2}{*}{\textbf{N}} 
    & \multicolumn{4}{c|}{\textbf{Arithmetics}}
    & \multicolumn{4}{c}{\textbf{Form.Log.}} 
    \\
    
    \cmidrule(lr){4-7} \cmidrule(lr){8-11} 
    
    & &
    & \textbf{R=0} & \textbf{R=1} & \textbf{R=2} & \textbf{Avg.}
    & \textbf{R=0} & \textbf{R=1} & \textbf{R=2} & \textbf{Avg.}
    \\
    
    \midrule

    \multirow{6}{*}{\rotatebox[origin=c]{90}{Qwen2.5-3B}}
    &
    SC & 10  
    & 94.5$\pm$4.0&	60.4$\pm$1.6	&34.6$\pm$5.8   & 63.1
    & 44.8$\pm$2.5&	35.3$\pm$2.9	&27.6$\pm$5.4   & 35.9
    \\

    &
    \cellcolor{green!15}RCS$_\text{uni}$ & \cellcolor{green!15}10
    & \cellcolor{green!15}94.3$\pm$3.7	&\cellcolor{green!15}59.5$\pm$4.1&\cellcolor{green!15}	34.0$\pm$11.0 & \cellcolor{green!15}62.6
    & \cellcolor{green!15}\textbf{46.0$\pm$2.3}	&\cellcolor{green!15}35.8$\pm$5.3	&\cellcolor{green!15}\textbf{28.8$\pm$6.0} & \cellcolor{green!15}\textbf{36.9}
    \\
    &
    \cellcolor{green!15}RCS$_\text{freq}$ & 10\cellcolor{green!15}
    & \cellcolor{green!15}\textbf{95.2$\pm$3.6}&\cellcolor{green!15}	\textbf{61.4$\pm$1.2}	&\cellcolor{green!15}\textbf{34.7$\pm$5.4} & \cellcolor{green!15}\textbf{63.8 }
    & \cellcolor{green!15}44.8$\pm$2.8&\cellcolor{green!15}	\textbf{36.1$\pm$3.3}&\cellcolor{green!15}	28.7$\pm$5.5 & \cellcolor{green!15}36.5
    \\

    \cmidrule(lr){2-11}
    &
    SC & 20  
    & 96.5$\pm$0.5&	24.8$\pm$3.0	&34.0$\pm$3.3 & 51.8
    & 42.3$\pm$0.5	&40.5$\pm$1.6	&35.7$\pm$1.6 & 39.5
    \\
    & 
    \cellcolor{green!15}RCS$_\text{uni}$ & 20\cellcolor{green!15}
    & \cellcolor{green!15}\textbf{97.7$\pm$1.0}	&\cellcolor{green!15}\textbf{27.3$\pm$3.2}	&\cellcolor{green!15}\textbf{34.5$\pm$5.1} & \cellcolor{green!15}\textbf{53.2} 
    & \cellcolor{green!15}\textbf{46.6$\pm$0.9}	&\cellcolor{green!15}\textbf{45.2$\pm$1.4}&	\cellcolor{green!15}\textbf{37.8$\pm$0.9} & \cellcolor{green!15}\textbf{43.2}
    \\
    &
    \cellcolor{green!15}RCS$_\text{freq}$ & 20\cellcolor{green!15}
    & \cellcolor{green!15}97.2$\pm$0.3	&\cellcolor{green!15}26.0$\pm$2.2&\cellcolor{green!15}	34.3$\pm$4.0 & \cellcolor{green!15}52.5
    & \cellcolor{green!15}42.6$\pm$0.5&	\cellcolor{green!15}42.1$\pm$0.8&\cellcolor{green!15}	35.7$\pm$1.6 & \cellcolor{green!15}40.1
    \\

    \midrule
    \multirow{6}{*}{\rotatebox[origin=c]{90}{Llama3.2-3B}}
    &
    SC & 10  
    & 80.7$\pm$2.0	&85.6$\pm$10.7&	75.8$\pm$5.9 & 80.7
    & 31.3$\pm$2.9&	28.4$\pm$7.8	&29.5$\pm$7.7 & 29.7
    \\
    &
    \cellcolor{green!15}RCS$_\text{uni}$ & 10\cellcolor{green!15}
    & \cellcolor{green!15}79.1$\pm$4.3	&\cellcolor{green!15} \textbf{85.8$\pm$9.1}&\cellcolor{green!15}	\textbf{76.2$\pm$4.6} & \cellcolor{green!15}80.4
    & 	\cellcolor{green!15}\textbf{33.2$\pm$0.5}&\cellcolor{green!15}	\textbf{28.7$\pm$8.0}	&\cellcolor{green!15}\textbf{29.9$\pm$6.4} & \cellcolor{green!15}\textbf{30.6}
    \\
    &
    \cellcolor{green!15}RCS$_\text{freq}$ & 10\cellcolor{green!15}
    & \cellcolor{green!15}\textbf{81.1$\pm$12.7}&	\cellcolor{green!15}85.6$\pm$10.0	&\cellcolor{green!15}\textbf{76.2$\pm$5.1} & \cellcolor{green!15}\textbf{81.0}
    & \cellcolor{green!15}32.1$\pm$2.9&	\cellcolor{green!15}28.6$\pm$8.3	&\cellcolor{green!15}29.5$\pm$16.6 & \cellcolor{green!15}30.1
    \\

    \cmidrule(lr){2-11}
    &
    SC & 20  
    & 99.0$\pm$0.5&	72.2$\pm$4.1&	41.5$\pm$2.2 & 70.9
    & 13.2$\pm$0.5&	3.2$\pm$0.8	&3.7$\pm$1.7 & 6.7
    \\
    &
    \cellcolor{green!15}RCS$_\text{uni}$ & 20\cellcolor{green!15}
    & \cellcolor{green!15}99.0$\pm$1.0	&\cellcolor{green!15}\textbf{76.5$\pm$4.3}	&\cellcolor{green!15}\textbf{43.3$\pm$1.5} & \cellcolor{green!15}\textbf{72.9}
    & \cellcolor{green!15}\textbf{21.4$\pm$2.4}&\cellcolor{green!15}	\textbf{9.0$\pm$1.2}&\cellcolor{green!15}	\textbf{7.1$\pm$2.4}&\cellcolor{green!15}\textbf{12.5}
    \\
    &
    \cellcolor{green!15}RCS$_\text{freq}$ & 20\cellcolor{green!15}
    & \cellcolor{green!15}\textbf{99.2$\pm$0.8}	&\cellcolor{green!15}73.8$\pm$3.8&\cellcolor{green!15}	42.5$\pm$2.6 & \cellcolor{green!15}71.8
    & \cellcolor{green!15}15.6$\pm$1.7	&\cellcolor{green!15}4.2$\pm$0.5&\cellcolor{green!15}	3.7$\pm$0.9 & \cellcolor{green!15}7.8
    \\

    \bottomrule
  \end{tabular}}
\end{table}
All RCS variants use \texttt{all-MiniLM-L6-v2}~\citep{reimers2019sentence}. We sample $N{=}10$ responses (also $N{=}5,20,40$) at $T{=}1$ via vLLM~\citep{kwon2023efficient}; results are mean$\pm$std over three seeds. 
Because we sample at $T{=}1$ and retain all outputs (including empty or degenerate ones), sampling-based selectors, including RCS, can fall below greedy decoding on some settings. Our evaluation isolates the selection step over a fixed candidate pool for each $N$ instead of reusing sample pools.
Additional details are in Appendix~\ref{app:details}.



\subsection{Main Results}

\paragraph{RCS Consistently Outperforms Baselines.}

We report selection accuracy in Table~\ref{tab:ranking}. RCS variants consistently achieve top performance across settings. RCS$_\text{uni}$, RCS$_\text{cosine}$, and RCS$_\text{prob}$ rank first overall, outperforming the best baseline (CE) by 2--6\%. 
Notably, RCS$_\text{cosine}$ outperforms SCW despite using the same pairwise cosine signals, highlighting the advantage of operating at the final-answer level and avoiding discrete grouping.
Similarly, RCS$_\text{freq}$ outperforms SC by avoiding hard tie-breaking via soft semantic weighting.
Among baselines, CE is the strongest competitor, while ModeX and SCW both underperform due to full-reasoning embedding collapse (see Section~\ref{sec:embedding-collapse}).
The gains are model-dependent rather than scale-dependent, with RCS benefiting most when answers exhibit greater lexical diversity.
Overall, a unified geometric framework with flexible weighting schemes is more expressive and robust than surface-form or trajectory-level aggregation.

\paragraph{RCS Remains Competitive Across Sampling Budgets.}\label{sec:scaling}

Figure~\ref{fig:scaling} shows performance as the number of sampled responses $N$ increases with only top methods are shown. We highlight $\text{RCS}_{\text{uni}}$ and $\text{RCS}_{\text{prob}}$, the best-performing variants from Table~\ref{tab:ranking} (full results in Appendix~\ref{app:sample-details}).
RCS matches or outperforms all baselines at every $N$, with gains up to +7\%. As $N$ increases and exact-match voting stabilizes on closed-form tasks (numeric or multiple-choice answers), RCS effectively reduces to voting, while its advantage persists on free-form answers such as GPQA, where correct responses remain lexically fragmented. The widening gaps are mainly driven by SCW, ModeX, which relies on full reasoning traces prone to collapse, and likelihood-based baselines degrading with $N$.

\paragraph{RCS Enhances Multi-Agent Debate.}

We evaluate RCS as a drop-in replacement for majority voting in multi-agent debate~\citep{du2023improving, choi2025debate} under fully black-box settings, using $N{=}10,20$ agents for $R{=}2$ rounds with Qwen2.5-3B and Llama3.2-3B. We report RCS$_\text{uni}$ and RCS$_\text{freq}$ (Table~\ref{tab:mad_qwen2.5-3b}), which are computationally efficient and do not require access to model internals.
Increasing the number of agents to $N=20$ degrades absolute accuracy across debate rounds, likely due to context truncation rather than the aggregation rule. We therefore focus on controlled comparisons using identical debate transcripts. RCS performs comparably to majority voting at $N=10$ and improves performance by 1–2\% on average at $N=20$, highlighting its robustness when aggregation becomes more challenging.

\paragraph{RCS Improves Under Fully Black-box Setting.}

\begin{table*}[t]
\centering
\caption{Performance under (a) black-box setting ($N{=}$10) and (b) complexity comparison.}
\label{tab:clean-blackbox-complexity}
\vspace{-5pt}

\resizebox{\linewidth}{!}{
\begin{tabular}{c c}


\begin{minipage}[t]{0.715\linewidth}
\centering
    (a)
\vspace{-0.5pt}

\label{tab:blackbox}
\resizebox{\linewidth}{!}{
\begin{tabular}{l|cccc|c}
\toprule
\textbf{Method} & \textbf{GPQA} & \textbf{MMLU-Pro} & \textbf{BBH-Date} & \textbf{BBH-Navigate} & {\textbf{Avg.}}\\
\midrule
SC  & 13.1 & 47.6 & 76.8 & 39.2 & 44.2\\
SCW & 12.6 & 47.6 & 76.2 & 39.7 & 44.0\\
\rowcolor{green!15}RCS$_\text{medoid}$ & \textbf{14.6} & \textbf{48.6} & 76.4 & 39.2 & 44.7\\
\rowcolor{green!15}RCS$_\text{uni}$   & 14.1 & \textbf{48.6} & \textbf{78.0} & 40.8 & \textbf{45.4}\\
\rowcolor{green!15}RCS$_\text{freq}$   & 13.6 & 47.6 & 77.6 & \textbf{41.2} & 45.0\\
\rowcolor{green!15}RCS$_\text{cosine}$   & 13.6 & 47.6 & 77.2 & 40.5 & 44.7\\
\bottomrule
\end{tabular}}
\end{minipage}

&

\begin{minipage}[t]{0.265\linewidth}
\centering




(b)
\vspace{-0.5pt}

\label{tab:complexity}
\resizebox{\linewidth}{!}{
\begin{tabular}{l c}
\toprule
\textbf{Method} & \textbf{Complexity} \\
\midrule
NLL, ANLL & $\mathcal{O}(NL)$ \\
SC & $\mathcal{O}(N)$ \\
SCW & $\mathcal{O}(N^2d)$ \\
CE & $\mathcal{O}(NL|V|)$ \\
ModeX & $\mathcal{O}(N^2L + N^3)$\\
\rowcolor{green!15}RCS$_\text{medoid/cosine}$ & $\mathcal{O}(N^2d)$ \\
\rowcolor{green!15}RCS$_\text{uni/freq}$ & $\mathcal{O}(Nd)$ \\
\rowcolor{green!15}RCS$_\text{prob}$ & $\mathcal{O}(N(d + L))$ \\
\bottomrule
\end{tabular}
}

\end{minipage}

\end{tabular}
}
\end{table*}

\begin{figure*}[ht]
\centering
\small
\fcolorbox{black}{white}{%
\begin{minipage}{0.98\linewidth}
\textbf{Question:} A spin-1/2 particle sits in a magnetic field along $+Z$; at $t{=}0$ the field is instantaneously switched to $+Y$, same magnitude $B$. What is the oscillation frequency of $\langle S_z\rangle$? \\[0.3em]
\textbf{Ground Truth}: \textbf{$\gamma B$} \\[0.6em]
\begin{tabular}{@{}>{\raggedright\arraybackslash}p{0.47\linewidth}>{\raggedright\arraybackslash}p{0.47\linewidth}@{}}
\textcolor{red}{\textbf{A1:}} \texttt{(gamma*B)/(2*pi)}. Let me verify... &
\textcolor{red}{\textbf{A2:}} \texttt{g*B/h} \\[0.3em]
\textcolor{red}{\textbf{A3:}} \texttt{$\omega$=2$\mu$B/$\hbar$} &
\textcolor{green!50!black}{\textbf{A4:}} \texttt{gamma*B} \\[0.3em]
\textcolor{green!50!black}{\textbf{A5:}} \texttt{gamma*B}. Wait, let me double-check... &
\textcolor{green!50!black}{\textbf{A6:}} \texttt{gamma*B} is the final... \\[0.3em]
\textcolor{red}{\textbf{A7:}} \texttt{2$\mu$B/\textbackslash hbar}. Hmm, actually... &
\textcolor{green!50!black}{\textbf{A8:}} \texttt{\textbackslash gamma B.} \\[0.3em]
\textcolor{green!50!black}{\textbf{A9:}} \texttt{\textbackslash gamma B}. So the frequency is... &
\textcolor{green!50!black}{\textbf{A10:}} \texttt{$\gamma$B}
\end{tabular}
\\[0.6em]

\begin{tabular}{p{0.47\linewidth}|p{0.47\linewidth}}
\centering
Self-Consistency fails to clearly distinguish between 10 verbatim-distinct strings (\textcolor{red}{\ding{55}}).
&
\centering
RCS selects \textcolor{green!50!black}{$\gamma B$} (\textcolor{green!50!black}{\ding{51}}), as A4--A6, A8--A10 embed near one center despite lexical fragmentation.
\end{tabular}

\end{minipage}%
}
\caption{Qualitative example showing RCS recovers correct answers more reliably than SC (GPQA, Gemma2-9B).}\label{fig:qualitative}
\end{figure*}


\begin{figure*}[ht]
    \centering
    \includegraphics[width=\linewidth]{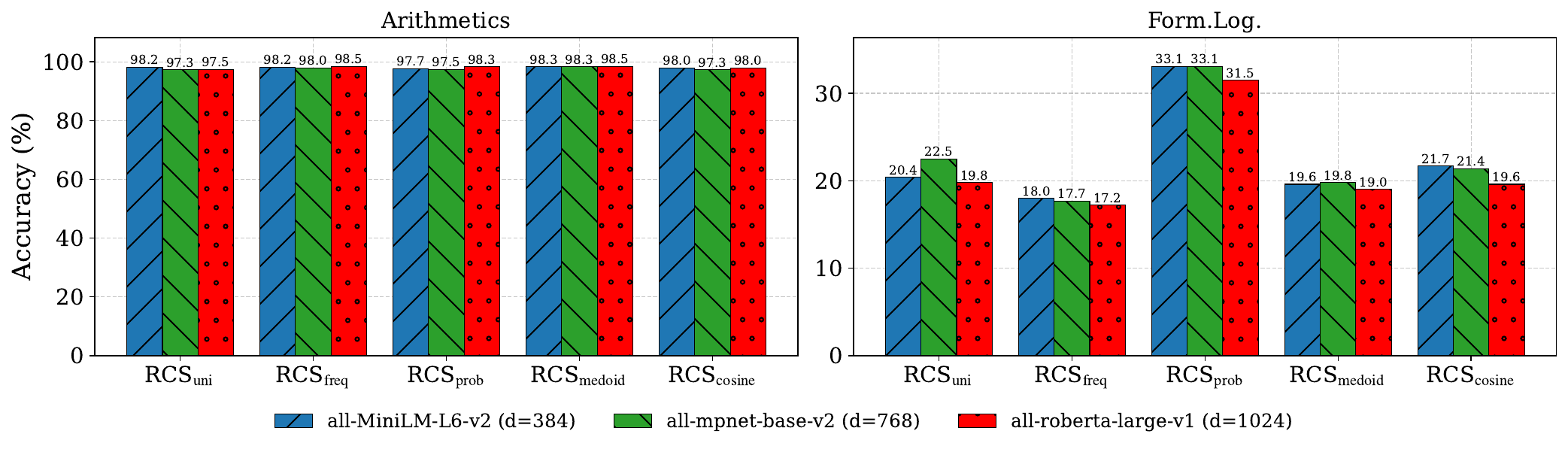}
    \caption{
    Effect of encoders on Llama3.2-3B. Results for Qwen2.5-3B are in Appendix Figure~\ref{fig:embed_qwen2.5-3b}.
    }
    \label{fig:embed_llama3.2-3b}
\end{figure*}

\begin{figure*}[ht]
    \centering
    \includegraphics[width=\linewidth]{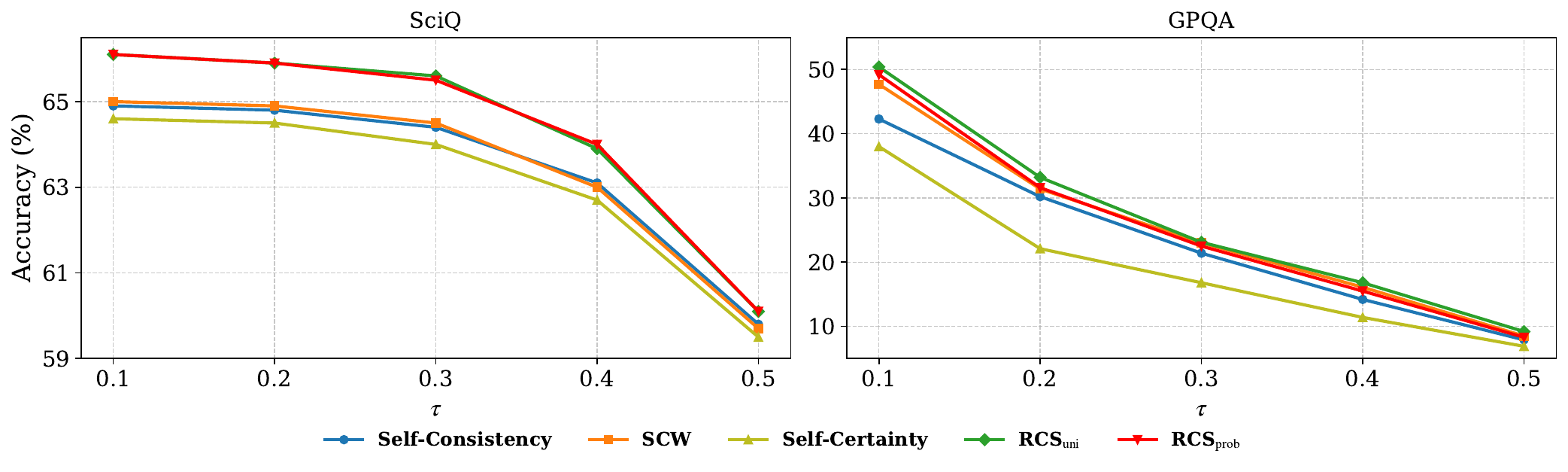}
    \caption{
    Effect of varying $\tau$ using Qwen2.5-3B. Results for Llama3.2-3B are in Appendix Figure~\ref{fig:threshold_llama3.2-3b}.
    }
    \label{fig:threshold_qwen2.5-3b}
\end{figure*}
\begin{figure*}[ht]
    \centering
    \includegraphics[width=\linewidth]{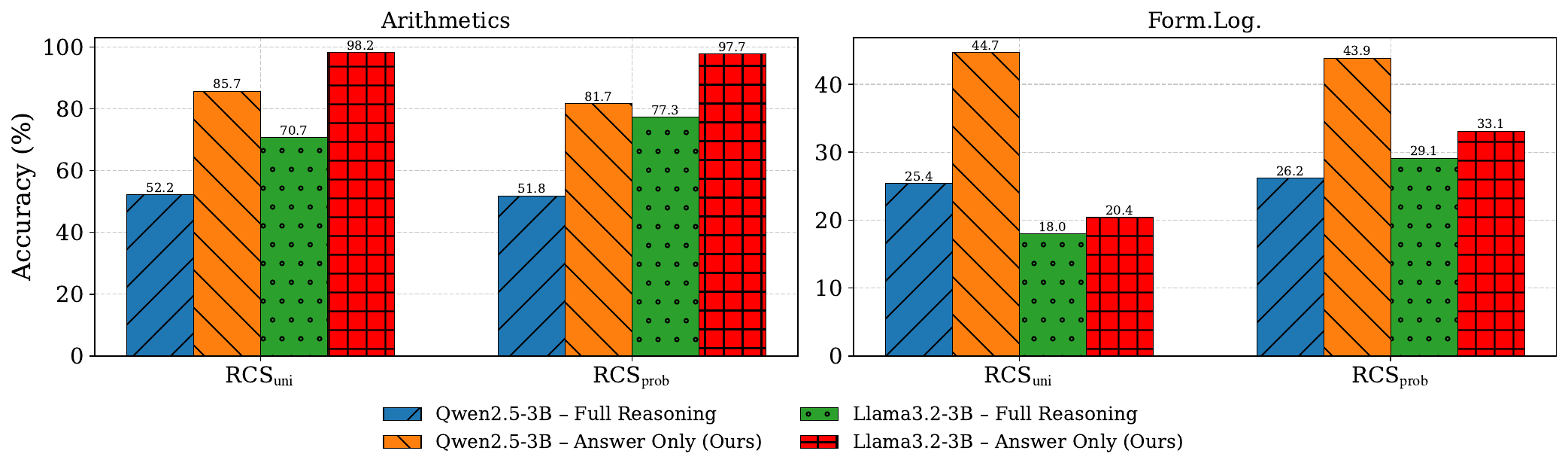}
    \caption{
    Effect of reasoning paths on Arithmetics and Formal Logic. 
    }
    \label{fig:full}
\end{figure*}

We further evaluate a fully black-box setting using Cohere (\textit{command-a-03-2025}) on GPQA, MMLU-Pro, and BigBenchHard (Table~\ref{tab:clean-blackbox-complexity}a). RCS outperforms traditional Self-Consistency, with RCS$_\text{uni}$ achieving the best performance, yielding +1\% average improvement. This highlights the advantage of RCS in capturing consistency beyond simple frequency-based selection.

\subsection{Ablation Study}\label{sec:ablation}


\paragraph{Complexity Analysis.}\label{sec:complexity}

We report computational complexity in Table~\ref{tab:clean-blackbox-complexity}b, where $d$, $L$, and $|V|$ denote embedding dimension, sequence length, and vocabulary size.
NLL and ANLL scale as $\mathcal{O}(NL)$ due to token-level scoring, while SC is $\mathcal{O}(N)$ since it only performs frequency counting over final answers. More expressive methods such as CE and ModeX capture richer semantic or structural dependencies but incur higher costs due to full token-level distributions or pairwise interactions.
In contrast, RCS variants (e.g., RCS$_\text{uni}$) avoid pairwise comparisons by estimating a semantic center and dispersion in $\mathcal{O}(Nd)$, achieving a favorable trade-off between efficiency and semantic expressiveness.

\paragraph{Qualitative Example.}\label{sec:qualitative}
We present a qualitative example from GPQA (Figure~\ref{fig:qualitative}). The correct answer ($\gamma B$) appears in 6/10 samples but splits into six verbatim-distinct strings across notation and trailing chatbot text, so Self-Consistency sees a full 10-way tie and falls back to a wrong tie-break, while RCS$_\text{uni}$'s centroid still lands inside the correct cluster.
On a harder boundary case from the same instance ($N{=}7$ (A1-A7), correct cluster size $k{=}3 < N/2$), real embeddings still place RCS$_\text{uni}$'s centroid nearest a correct candidate (cosine distance 0.498 vs.\ 0.609--0.899 for the wrong ones), even though $k<N/2$. Here the correct cluster is a semantic plurality rather than a majority, as the four wrong answers split into three distinct clusters, a multi-cluster case not covered by Proposition~\ref{prop:finite}'s two-cluster analysis. Recovering a genuine semantic minority instead requires an informative weighting such as RCS$_\text{prob}$ (Figure~\ref{fig:overview}b).

\paragraph{Effect Of Embedding Models.}

In Figure~\ref{fig:embed_llama3.2-3b}, we compare three sentence transformers of increasing capacity.
Results show minimal variation across models. This stability is especially evident on Arithmetics, where answers are numeric. On Form.Log., we observe a slight drop with \texttt{all-roberta-large-v1}, particularly for the probability-based variant, suggesting that higher embedding capacity does not necessarily improve semantic alignment for this task.

\paragraph{Effect Of Correctness Threshold.}

Figure~\ref{fig:threshold_qwen2.5-3b} reports performance on SciQ and GPQA under varying correctness thresholds. We focus on \textsc{RCS}$_{\text{uni}}$ and \textsc{RCS}$_{\text{prob}}$, which consistently achieve the best results as the ROUGE-F1 threshold increases from 0.1 to 0.5, particularly within the common range of 0.2--0.4. This demonstrates RCS's robustness to threshold selection.


\paragraph{Effect of Full-Trajectory Embeddings.}\label{sec:embedding-collapse}
A natural extension of RCS is to use full generation trajectories instead of final-answer embeddings. 
However, this leads to an \textit{embedding collapse} effect, where candidates become less distinguishable due to shared prefixes and redundant reasoning.
We compare this variant with final-answer embeddings (Figure~\ref{fig:full}) and observe consistent performance drops.
This collapse issue also affects methods that operate on full trajectories, such as ModeX and SCW, which cluster or group complete reasoning traces based on embedding similarity. 
This suggests that intermediate tokens introduce noise and weaken discriminative signals, while final answers provide a more compact and task-relevant representation. 
RCS avoids both issues by restricting embeddings to the extracted final answer and ranking all candidates continuously.



\section{Conclusion}
We introduce Radial Consensus Score (RCS), an efficient method for reliable best-of-$N$ selection in LLMs. 
RCS ranks candidates by their distance to a weighted Fr\'echet mean of answer embeddings, where the weighting scheme (uniform, frequency, cosine-similarity, or probability) determines how consensus is defined.
RCS matches or outperforms strong baselines across diverse benchmarks and sampling budgets, with the largest gains when correct answers are lexically fragmented, while being efficient, black-box compatible, and applicable as a drop-in replacement for majority voting in multi-agent debate. 
This highlights geometric consensus as a simple and scalable principle for reliable answer selection.




\section*{Limitations}\label{app:limitation}
While efficient, RCS relies on embedding quality and assumes that correct answers cluster around a semantic centroid, which may cause semantically isolated correct answers to be overlooked when supported by only a small minority. This assumption may be less effective with diverse valid answers or tightly clustered hallucinations from shared reasoning flaws. 
Moreover, RCS assumes a fixed budget of N samples per question; pairing it with an uncertainty-aware allocation scheme such as~\citep{nguyen2026uncertainty} is a natural extension.
Improving embedding robustness and better handling such diversity are promising directions for future work.

\section*{Societal Impacts}\label{app:impact}

RCS can positively impact applications by improving the reliability of best-of-$N$ selection in LLMs, leading to more consistent and accurate outputs. However, its effectiveness depends on the quality of embedding representations, which may inherit or amplify biases from the underlying model. Therefore, caution is needed when applying it in high-stakes or safety-critical settings.

\section*{Acknowledgement}
This research was funded partially by the Australian Government through the Australian Research Council. Dr. Hung Le is the recipient of an Australian Research Council Discovery Early Career Researcher Award (project number DE250100355) funded by the Australian Government.

\bibliographystyle{plain} 
\bibliography{references}

@article{wang2022self,
  title={Self-consistency improves chain of thought reasoning in language models},
  author={Wang, Xuezhi and Wei, Jason and Schuurmans, Dale and Le, Quoc and Chi, Ed and Narang, Sharan and Chowdhery, Aakanksha and Zhou, Denny},
  journal={arXiv preprint arXiv:2203.11171},
  year={2022}
}

@article{brown2020language,
  title={Language models are few-shot learners},
  author={Brown, Tom and Mann, Benjamin and Ryder, Nick and Subbiah, Melanie and Kaplan, Jared D and Dhariwal, Prafulla and Neelakantan, Arvind and Shyam, Pranav and Sastry, Girish and Askell, Amanda and others},
  journal={Advances in neural information processing systems},
  volume={33},
  pages={1877--1901},
  year={2020}
}

@inproceedings{du2023improving,
  title={Improving factuality and reasoning in language models through multiagent debate},
  author={Du, Yilun and Li, Shuang and Torralba, Antonio and Tenenbaum, Joshua B and Mordatch, Igor},
  booktitle={Forty-first International Conference on Machine Learning},
  year={2023}
}

@article{yang2025qwen25,
  title={{Qwen2.5} technical report},
  author={Qwen Team},
  journal={arXiv preprint arXiv:2412.15115},
  year={2025}
}

@article{reimers2019sentence,
  title={Sentence-bert: Sentence embeddings using siamese bert-networks},
  author={Reimers, Nils and Gurevych, Iryna},
  journal={arXiv preprint arXiv:1908.10084},
  year={2019}
}

@article{wei2022chain,
  title={Chain-of-thought prompting elicits reasoning in large language models},
  author={Wei, Jason and Wang, Xuezhi and Schuurmans, Dale and Bosma, Maarten and Xia, Fei and Chi, Ed and Le, Quoc V and Zhou, Denny and others},
  journal={Advances in neural information processing systems},
  volume={35},
  pages={24824--24837},
  year={2022}
}

@article{grattafiori2024llama,
  title={The llama 3 herd of models},
  author={Grattafiori, Aaron and Dubey, Abhimanyu and Jauhri, Abhinav and Pandey, Abhinav and Kadian, Abhishek and Al-Dahle, Ahmad and Letman, Aiesha and Mathur, Akhil and Schelten, Alan and Vaughan, Alex and others},
  journal={arXiv preprint arXiv:2407.21783},
  year={2024}
}

@inproceedings{guerreiro2022looking,
  title={Looking for a Needle in a Haystack: A Comprehensive Study of Hallucinations in Neural Machine Translation},
  author={Guerreiro, Nuno M and Voita, Elena and Martins, Andr{\'e} FT},
  booktitle={Proceedings of the 17th Conference of the European Chapter of the Association for Computational Linguistics},
  pages={1059--1075},
  year={2023}
}

@article{farquhar2024detecting,
  title={Detecting hallucinations in large language models using semantic entropy},
  author={Farquhar, Sebastian and Kossen, Jannik and Kuhn, Lorenz and Gal, Yarin},
  journal={Nature},
  volume={630},
  number={8017},
  pages={625--630},
  year={2024},
  publisher={Nature Publishing Group UK London}
}

@inproceedings{manakul2023selfcheckgpt,
  title={SelfCheckGPT: Zero-Resource Black-Box Hallucination Detection for Generative Large Language Models},
  author={Manakul, Potsawee and Liusie, Adian and Gales, Mark},
  booktitle={Proceedings of the 2023 Conference on Empirical Methods in Natural Language Processing},
  pages={9004--9017},
  year={2023}
}

@article{lin2023generating,
  title={Generating with confidence: Uncertainty quantification for black-box large language models},
  author={Lin, Zhen and Trivedi, Shubhendu and Sun, Jimeng},
  journal={arXiv preprint arXiv:2305.19187},
  year={2023}
}

@inproceedings{kuhn2023semantic,
  title={Semantic Uncertainty: Linguistic Invariances for Uncertainty Estimation in Natural Language Generation},
  author={Kuhn, Lorenz and Gal, Yarin and Farquhar, Sebastian},
  booktitle={The Eleventh International Conference on Learning Representations},
  year={2023}
}

@article{nikitin2024kernel,
  title={Kernel language entropy: Fine-grained uncertainty quantification for LLMs from semantic similarities},
  author={Nikitin, Alexander and Kossen, Jannik and Gal, Yarin and Marttinen, Pekka},
  journal={Advances in Neural Information Processing Systems},
  volume={37},
  pages={8901--8929},
  year={2024}
}

@article{welbl2017crowdsourcing,
  title={Crowdsourcing multiple choice science questions},
  author={Welbl, Johannes and Liu, Nelson F and Gardner, Matt},
  journal={arXiv preprint arXiv:1707.06209},
  year={2017}
}

@article{team2024gemma,
  title={{Gemma}: Open models based on {Gemini} research and technology},
  author={Mesnard, Thomas and Hardin, Cassidy and Dadashi, Robert and Bhupatiraju, Surya and Pathak, Shreya and Sifre, Laurent and Rivi{\`e}re, Morgane and Kale, Mihir Sanjay and Love, Juliette and others},
  journal={arXiv preprint arXiv:2403.08295},
  year={2024}
}

@article{huang2025survey,
  title={A survey on hallucination in large language models: Principles, taxonomy, challenges, and open questions},
  author={Huang, Lei and Yu, Weijiang and Ma, Weitao and Zhong, Weihong and Feng, Zhangyin and Wang, Haotian and Chen, Qianglong and Peng, Weihua and Feng, Xiaocheng and Qin, Bing and others},
  journal={ACM Transactions on Information Systems},
  volume={43},
  number={2},
  pages={1--55},
  year={2025},
  publisher={ACM New York, NY}
}

@inproceedings{nguyen2025probabilities,
  title={Probabilities Are All You Need: A Probability-Only Approach to Uncertainty Estimation in Large Language Models},
  author={Nguyen, Manh and Gupta, Sunil and Le, Hung},
  booktitle={Proceedings of the AAAI Conference on Artificial Intelligence},
  year={2026},
  pages={32546--32554}
}

@inproceedings{rein2024gpqa,
  title={{GPQA}: A graduate-level {Google}-proof {Q\&A} benchmark},
  author={Rein, David and Hou, Betty Li and Stickland, Asa Cooper and Petty, Jackson and Pang, Richard Yuanzhe and Dirani, Julien and Michael, Julian and Bowman, Samuel R},
  booktitle={First Conference on Language Modeling},
  year={2024}
}

@inproceedings{kwon2023efficient,
  title={Efficient Memory Management for Large Language Model Serving with PagedAttention},
  author={Woosuk Kwon and Zhuohan Li and Siyuan Zhuang and Ying Sheng and Lianmin Zheng and Cody Hao Yu and Joseph E. Gonzalez and Hao Zhang and Ion Stoica},
  booktitle={Proceedings of the ACM SIGOPS 29th Symposium on Operating Systems Principles},
  year={2023}
}

@article{kang2025scalable,
  title={Scalable best-of-n selection for large language models via self-certainty},
  author={Kang, Zhewei and Zhao, Xuandong and Song, Dawn},
  journal={arXiv preprint arXiv:2502.18581},
  year={2025}
}

@inproceedings{wang2024integrate,
  title={Integrate the Essence and Eliminate the Dross: Fine-Grained Self-Consistency for Free-Form Language Generation},
  author={Wang, Xinglin and Li, Yiwei and Feng, Shaoxiong and Yuan, Peiwen and Pan, Boyuan and Wang, Heda and Hu, Yao and Li, Kan},
  booktitle={Proceedings of the 62nd Annual Meeting of the Association for Computational Linguistics (Volume 1: Long Papers)},
  pages={11782--11794},
  year={2024}
}

@article{chen2023universal,
  title={Universal self-consistency for large language model generation},
  author={Chen, Xinyun and Aksitov, Renat and Alon, Uri and Ren, Jie and Xiao, Kefan and Yin, Pengcheng and Prakash, Sushant and Sutton, Charles and Wang, Xuezhi and Zhou, Denny},
  journal={arXiv preprint arXiv:2311.17311},
  year={2023}
}

@article{choi2025debate,
  title={Debate or vote: Which yields better decisions in multi-agent large language models?},
  author={Choi, Hyeong Kyu and Zhu, Xiaojin and Li, Sharon},
  journal={arXiv preprint arXiv:2508.17536},
  year={2025}
}

@article{cobbe2021training,
  title={Training verifiers to solve math word problems},
  author={Cobbe, Karl and Kosaraju, Vineet and Bavarian, Mohammad and Chen, Mark and Jun, Heewoo and Kaiser, Lukasz and Plappert, Matthias and Tworek, Jerry and Hilton, Jacob and Nakano, Reiichiro and others},
  journal={arXiv preprint arXiv:2110.14168},
  year={2021}
}

@article{hendrycks2020aligning,
  title={Aligning ai with shared human values},
  author={Hendrycks, Dan and Burns, Collin and Basart, Steven and Critch, Andrew and Li, Jerry and Song, Dawn and Steinhardt, Jacob},
  journal={arXiv preprint arXiv:2008.02275},
  year={2020}
}

@inproceedings{wang2025ranked,
  title={Ranked voting based self-consistency of large language models},
  author={Wang, Weiqin and Wang, Yile and Huang, Hui},
  booktitle={Findings of the Association for Computational Linguistics: ACL 2025},
  pages={14410--14426},
  year={2025}
}

@inproceedings{wan2025reasoning,
  title={Reasoning aware self-consistency: Leveraging reasoning paths for efficient llm sampling},
  author={Wan, Guangya and Wu, Yuqi and Chen, Jie and Li, Sheng},
  booktitle={Proceedings of the 2025 Conference of the Nations of the Americas Chapter of the Association for Computational Linguistics: Human Language Technologies (Volume 1: Long Papers)},
  pages={3613--3635},
  year={2025}
}

@article{gui2024bonbon,
  title={Bonbon alignment for large language models and the sweetness of best-of-n sampling},
  author={Gui, Lin and G{\^a}rbacea, Cristina and Veitch, Victor},
  journal={Advances in Neural Information Processing Systems},
  volume={37},
  pages={2851--2885},
  year={2024}
}

@article{cordero2025certified,
  title={Certified self-consistency: Statistical guarantees and test-time training for reliable reasoning in LLMs},
  author={Cordero-Encinar, Paula and Duncan, Andrew B},
  journal={arXiv preprint arXiv:2510.17472},
  year={2025}
}

@article{nguyen2026hear,
  title={Hear Both Sides: Efficient Multi-Agent Debate via Diversity-Aware Message Retention},
  author={Nguyen, Manh and Nguyen, Anh and Nguyen, Dung and Venkatesh, Svetha and Le, Hung},
  journal={arXiv preprint arXiv:2603.20640},
  year={2026}
}

@article{wang2024mmlu,
  title={{MMLU-Pro}: A more robust and challenging multi-task language understanding benchmark},
  author={Wang, Yubo and Ma, Xueguang and Zhang, Ge and Ni, Yuansheng and Chandra, Abhranil and Guo, Shiguang and Ren, Weiming and Arulraj, Aaran and He, Xuan and Jiang, Ziyan and others},
  journal={Advances in Neural Information Processing Systems},
  volume={37},
  pages={95266--95290},
  year={2024}
}

@article{fletcher,
  author={Fletcher, P.T. and Conglin Lu and Pizer, S.M. and Sarang Joshi},
  journal={IEEE Transactions on Medical Imaging}, 
  title={Principal geodesic analysis for the study of nonlinear statistics of shape}, 
  year={2004},
  volume={23},
  number={8},
  pages={995-1005},
  doi={10.1109/TMI.2004.831793}
  }

@inproceedings{wang-etal-2024-math,
    title = "Math-Shepherd: Verify and Reinforce {LLM}s Step-by-step without Human Annotations",
    author = "Wang, Peiyi  and
      Li, Lei  and
      Shao, Zhihong  and
      Xu, Runxin  and
      Dai, Damai  and
      Li, Yifei  and
      Chen, Deli  and
      Wu, Yu  and
      Sui, Zhifang",
    editor = "Ku, Lun-Wei  and
      Martins, Andre  and
      Srikumar, Vivek",
    booktitle = "Proceedings of the 62nd Annual Meeting of the Association for Computational Linguistics (Volume 1: Long Papers)",
    month = aug,
    year = "2024",
    address = "Bangkok, Thailand",
    publisher = "Association for Computational Linguistics",
    url = "https://aclanthology.org/2024.acl-long.510/",
    doi = "10.18653/v1/2024.acl-long.510",
    pages = "9426--9439",
}

@inproceedings{lightman2023let,
  title={Let's verify step by step},
  author={Lightman, Hunter and Kosaraju, Vineet and Burda, Yuri and Edwards, Harrison and Baker, Bowen and Lee, Teddy and Leike, Jan and Schulman, John and Sutskever, Ilya and Cobbe, Karl},
  booktitle={The twelfth international conference on learning representations},
  year={2023}
}

@article{di2025best,
  title={Best-of-Majority: Minimax-Optimal Strategy for Pass@ $ k $ Inference Scaling},
  author={Di, Qiwei and Ji, Kaixuan and Li, Xuheng and Zhao, Heyang and Gu, Quanquan},
  journal={arXiv preprint arXiv:2510.03199},
  year={2025}
}

@article{ai2025beyond,
  title={Beyond majority voting: LLM aggregation by leveraging higher-order information},
  author={Ai, Rui and Pan, Yuqi and Simchi-Levi, David and Tambe, Milind and Xu, Haifeng},
  journal={arXiv preprint arXiv:2510.01499},
  year={2025}
}

@inproceedings{choi2026modex,
  title     = {ModeX: Evaluator-Free Best-of-N Selection for Open-Ended Generation},
  author    = {Choi, Hyeong Kyu and Li, Sharon},
  booktitle = {Proceedings of the 64th Annual Meeting of the Association for Computational Linguistics},
  year      = {2026},
}

@inproceedings{suzgun2023challenging,
  title={Challenging {BIG-Bench} tasks and whether chain-of-thought can solve them},
  author={Suzgun, Mirac and Scales, Nathan and Sch{\"a}rli, Nathanael and Gehrmann, Sebastian and Tay, Yi and Chung, Hyung Won and Chowdhery, Aakanksha and Le, Quoc and Chi, Ed and Zhou, Denny and others},
  booktitle={Findings of the Association for Computational Linguistics: ACL 2023},
  pages={13003--13051},
  year={2023}
}

@inproceedings{deguchi2024centroid,
    title = "Centroid-Based Efficient Minimum {B}ayes Risk Decoding",
    author = "Deguchi, Hiroyuki  and
      Sakai, Yusuke  and
      Kamigaito, Hidetaka  and
      Watanabe, Taro  and
      Tanaka, Hideki  and
      Utiyama, Masao",
    editor = "Ku, Lun-Wei  and
      Martins, Andre  and
      Srikumar, Vivek",
    booktitle = "Findings of the Association for Computational Linguistics: ACL 2024",
    month = aug,
    year = "2024",
    address = "Bangkok, Thailand",
    publisher = "Association for Computational Linguistics",
    url = "https://aclanthology.org/2024.findings-acl.654/",
    doi = "10.18653/v1/2024.findings-acl.654",
    pages = "11009--11018"
}

@inproceedings{chen2024latent,
  title={Latent self-consistency for reliable majority-set selection in short-and long-answer reasoning},
  author={Oh, Jungsuk and Lee, Jay-Yoon},
  booktitle={Proceedings of the AAAI Conference on Artificial Intelligence},
  volume={40},
  number={38},
  pages={32591--32599},
  year={2026}
}

@article{jain2023sc,
  title={Semantic self-consistency: Enhancing language model reasoning via semantic weighting},
  author={Knappe, Tim and Li, Ryan and Chauhan, Ayush and Chhua, Kaylee and Zhu, Kevin and O'Brien, Sean},
  journal={arXiv preprint arXiv:2410.07839},
  year={2024}
}

@article{hoeffding1963,
 ISSN = {01621459, 1537274X},
 author = {Wassily Hoeffding},
 journal = {Journal of the American Statistical Association},
 number = {301},
 pages = {13--30},
 publisher = {[American Statistical Association, Taylor & Francis, Ltd.]},
 title = {Probability Inequalities for Sums of Bounded Random Variables},
 urldate = {2026-07-25},
 volume = {58},
 year = {1963}
}

@article{nguyen2026uncertainty,
   title={Uncertainty-Aware Budget Allocation for Adaptive Test-Time Reasoning},
   author={Nguyen, Manh and Gupta, Sunil and Le, Hung},
   journal={arXiv preprint arXiv:2605.26849},
   year={2026}
}

@article{maryanskyy2026agents,
  title={When agents disagree: The selection bottleneck in multi-agent LLM pipelines},
  author={Maryanskyy, Artem and Budnikov, Dmitry and Kaliyev, Alibek T},
  journal={Applied Sciences},
  volume={16},
  number={10},
  pages={4914},
  year={2026},
  publisher={MDPI}
}

\clearpage
\appendix
\section{Appendix}\label{sec:app}

\begin{algorithm}[h]
\caption{Best-of-$N$ Selection with Radial Consensus Score (RCS)}
\label{alg:rds}
\begin{algorithmic}[1]
\Require Answers $\{a_i\}_{i=1}^N$, distribution $P$, embedding model $\mathbf{E}$, mode $\in \{\text{continuous}, \text{discrete}\}$
\Ensure Selected index $i^*$

\Statex \textbf{Embeddings:} $\mathbf{u}_i = E(a_i)$ \quad for $i=1,\dots,N$

\If{mode = continuous}
    \State $\mathbf{c} \leftarrow \sum_{i=1}^N p_i \mathbf{u}_i$
\Else
    \State $\mathbf{c} \leftarrow \arg\min_{\mathbf{u}_j} \sum_{i=1}^N p_i \|\mathbf{u}_i - \mathbf{u}_j\|_2^2$
\EndIf

\State $i^* \leftarrow \arg\min_i \|\mathbf{u}_i - \mathbf{c}\|_2$\\
\Return $i^*$
\end{algorithmic}
\end{algorithm}

\subsection{Proof of Geometric Center Estimation}\label{app:proof_center}

Expanding the objective, we have:
\begin{align}
\sum_{i=1}^N p_i \|\mathbf{u}_i - \mathbf{z}\|_2^2
&= \sum_{i=1}^N p_i \left( \|\mathbf{u}_i\|_2^2 - 2 \mathbf{u}_i^\top \mathbf{z} + \|\mathbf{z}\|_2^2 \right) \\
&= \sum_{i=1}^N p_i \|\mathbf{u}_i\|_2^2 - 2 \mathbf{z}^\top \sum_{i=1}^N p_i \mathbf{u}_i + \|\mathbf{z}\|_2^2.
\end{align}

Taking derivative with respect to $\mathbf{z}$ and setting it to zero:
\begin{align}
-2 \sum_{i=1}^N p_i \mathbf{u}_i + 2 \mathbf{z} = 0,
\end{align}
which yields:
\begin{align}
\mathbf{z} = \sum_{i=1}^N p_i \mathbf{u}_i.
\end{align}

Since the objective is strictly convex in $\mathbf{z}$, this solution is unique.


\subsection{Proof of Proposition~\ref{prop:asymptotic}}
\label{app:proof_prop1}

\paragraph{Setup.}
The candidates $\mathbf{u}_1,\dots,\mathbf{u}_N$ are i.i.d.\ on $\Sph$ with
$\E[\mathbf{u}]=\rho\mathbf{c}^*$ and $\rho>0$ (Assumption~\eqref{eq:assumption}).
Recall that $\mathbf{c}^*:=\E[\mathbf{u}]/\|\E[\mathbf{u}]\|_2$ is a unit vector by
construction, so $\mathbf{c}^*\in\Sph$. Because $\|\mathbf{u}_i\|_2=1$, the norm
$\|\mathbf{u}\|_2$ is bounded, hence $\E\|\mathbf{u}\|_2\le 1<\infty$; the finite
first moment required by the strong law of large numbers (SLLN) holds automatically,
with no further integrability assumption. By the selection
identity~\eqref{eq:selection-identity} it suffices to show
$\mathbf{c}(P)/\|\mathbf{c}(P)\|_2 \xrightarrow{a.s.}\mathbf{c}^*$, since the ranking
$\argmax_i\langle\mathbf{u}_i,\cdot\rangle$ is continuous in the direction of its
argument and invariant to its positive scaling.

\paragraph{Geometric group (uniform, frequency).}
Here the weights do not depend on $\{\mathbf{u}_i\}$. For uniform weighting
$p_i=1/N$,
\begin{align}
\mathbf{c}(P)=\frac1N\sum_{i=1}^N\mathbf{u}_i
\;\xrightarrow{a.s.}\;\E[\mathbf{u}]=\rho\mathbf{c}^*
\end{align}
by the SLLN, so $\mathbf{c}(P)/\|\mathbf{c}(P)\|_2\to\rho\mathbf{c}^*/\rho=\mathbf{c}^*$
almost surely. For frequency weighting, $p_i=f_i/\sum_j f_j$ with $f_i$ the lexical
count of answer $i$. Writing
$\mathbf{c}(P)=\big(\tfrac1N\sum_i f_i\mathbf{u}_i\big)\big/\big(\tfrac1N\sum_i f_i\big)$,
the counts $f_i$ depend only on the discrete answers, not on the embedding noise
given the answers; numerator and denominator each converge a.s.\ by the SLLN to
$\E[f\mathbf{u}]$ and $\E[f]>0$. Since counts are invariant across paraphrases of the
same semantic answer, $\E[f\mathbf{u}]$ is a positive multiple of $\mathbf{c}^*$, and
the ratio is directionally consistent.

\paragraph{Endogenous variant (cosine) and exogenous variant (probability).}
Let $\bar{\mathbf{u}}:=\tfrac1N\sum_{j=1}^N\mathbf{u}_j$ denote the empirical mean
embedding. Both variants take the self-normalized form
\begin{align}
\label{eq:selfnorm}
\mathbf{c}(P)
=\frac{\tfrac1N\sum_i w(\mathbf{u}_i)\,\mathbf{u}_i}{\tfrac1N\sum_i w(\mathbf{u}_i)},
\end{align}
where the weight is $w(\mathbf{u})=\langle\mathbf{u},\bar{\mathbf{u}}\rangle$ for
the cosine variant (the average cosine similarity of $\mathbf{u}$ to the other
candidates, as in Table~\ref{tab:rds_variants}), and $w(\mathbf{u})=\Pr(y|x)$
for the probability variant. Both weights are bounded and continuous on the compact
sphere. By the SLLN the numerator converges a.s.\ to $\E[w(\mathbf{u})\mathbf{u}]$
and the denominator to $\E[w(\mathbf{u})]>0$; by the continuous mapping theorem,
\begin{align}
\label{eq:m-limit}
\mathbf{c}(P)\;\xrightarrow{a.s.}\;
\mathbf{m}:=\frac{\E[w(\mathbf{u})\mathbf{u}]}{\E[w(\mathbf{u})]}.
\end{align}
(For the cosine variant, $\bar{\mathbf{u}}\xrightarrow{a.s.}\rho\mathbf{c}^*$ by the
geometric case above, so the effective weight is
$w(\mathbf{u})\to\rho\langle\mathbf{u},\mathbf{c}^*\rangle$, a function of
$\langle\mathbf{u},\mathbf{c}^*\rangle$ alone.)

It remains to show $\mathbf{m}$ points along $\mathbf{c}^*$. Assume $\mathcal D$ is
\emph{rotationally symmetric about $\mathbf{c}^*$}: its density depends on
$\mathbf{u}$ only through $t:=\langle\mathbf{u},\mathbf{c}^*\rangle$. Every point on
the sphere decomposes exactly as
\begin{align}
\label{eq:decomp}
\mathbf{u}=t\,\mathbf{c}^*+\sqrt{1-t^2}\,\mathbf{s},
\qquad
t=\langle\mathbf{u},\mathbf{c}^*\rangle\in[-1,1],\quad
\mathbf{s}\in\Sph\cap(\mathbf{c}^*)^\perp ,
\end{align}
which is an identity for any $\mathbf{u}$, not an assumption. Under rotational
symmetry about $\mathbf{c}^*$, the conditional law of $\mathbf{s}$ given $t$ is the
uniform distribution on the unit sphere of $(\mathbf{c}^*)^\perp$, which is invariant
under every rotation fixing $\mathbf{c}^*$; the only fixed point of all such rotations
is the origin, so
\begin{align}
\label{eq:Es-zero}
\E[\mathbf{s}|t]=\mathbf 0,\qquad\text{hence}\qquad \E[\mathbf{s}]=\mathbf 0 .
\end{align}
Suppose the weight depends on $\mathbf{u}$ only through $t$, that is there exists a
function $\tilde w:[-1,1]\to\R_{\ge0}$ with $w(\mathbf{u})=\tilde w(t)$. This holds
for cosine with $\tilde w(t)=\rho t$ (from the limit above), and for the calibrated
probability model with $\tilde w=g$, where $g$ is the increasing likelihood profile
introduced below. Substituting~\eqref{eq:decomp} and using~\eqref{eq:Es-zero},
\begin{align}
\E[w(\mathbf{u})\mathbf{u}]
=\underbrace{\E[\tilde w(t)\,t]}_{>0}\,\mathbf{c}^*
+\E\!\big[\tilde w(t)\sqrt{1-t^2}\,\E[\mathbf{s}|t]\big]
=\E[\tilde w(t)\,t]\,\mathbf{c}^*,
\end{align}
where $\E[\tilde w(t)\,t]>0$ because $\tilde w(t)\ge0$ is a weight whose mass lies at
$t>0$ (near $\mathbf{c}^*$, since $\rho>0$). Therefore
$\mathbf{m}=\big(\E[\tilde w(t)\,t]/\E[\tilde w(t)]\big)\mathbf{c}^*$ is a positive
multiple of $\mathbf{c}^*$, and
$\mathbf{c}(P)/\|\mathbf{c}(P)\|_2\xrightarrow{a.s.}\mathbf{c}^*$.
 
\paragraph{On the probability weight.}
The step $w(\mathbf{u})=\tilde w(t)$ for the probability variant is a
\emph{calibration assumption}, not a fact: we assume the likelihood factors through
alignment to the semantic center, $\Pr(y|x)=g(t)$ for some increasing
$g:[-1,1]\to\R_{\ge0}$, so that answers closer to $\mathbf{c}^*$ (larger $t$) receive
larger weight. When calibration fails, $\mathbf{m}$ need not point along
$\mathbf{c}^*$; we therefore treat the probability variant's guarantee as conditional
on this assumption.
 
\paragraph{Remark (necessity of symmetry).}
Equation~\eqref{eq:Es-zero} is the only place rotational symmetry is used: it forces
the orthogonal component of $\E[w(\mathbf{u})\mathbf{u}]$ to vanish, leaving
$\mathbf{m}$ parallel to $\mathbf{c}^*$. If $\mathcal D$ is skewed off axis, then
$\E[\mathbf{s}]\neq\mathbf 0$, the orthogonal term no longer cancels, and $\mathbf{m}$
tilts away from $\mathbf{c}^*$, so the cosine and probability variants lose their
directional guarantee. The geometric group is unaffected, as its weights never
depend on $\mathbf{u}$.

\paragraph{Conclusion: probability of correct selection.}
Since $\mathbf{c}(P)/\|\mathbf{c}(P)\|_2\xrightarrow{a.s.}\mathbf{c}^*$ in all four
variants, the selected candidate satisfies
\begin{align}
\hat i=\argmax_i\langle\mathbf{u}_i,\mathbf{c}(P)\rangle
\xrightarrow{a.s.}\argmax_i\langle\mathbf{u}_i,\mathbf{c}^*\rangle,
\end{align}
i.e.\ RCS selects the candidate best aligned with the semantic center, and
\begin{align}
\Pr(\text{RCS selects the semantically correct answer})\to 1
\qquad\text{as }N\to\infty .
\end{align}
Exact-match voting, by contrast, operates on surface strings. When the correct
semantic cluster $a^*$ is split across $L\ge2$ surface forms
(e.g.\ ``Canberra'', ``It's Canberra'', ``Canberra ACT'') with no single form
holding a plurality, a compact wrong cluster with one surface form can win. With
$L\ge2$ fragments of $a^*$, no fragment attains plurality and voting selects $a^*$
with probability at most $1/L$, bounded away from $1$. Thus, as $N\to\infty$, RCS
attains probability $1$ while voting remains capped at $1/L$: RCS is asymptotically
superior whenever surface fragmentation occurs. \qed

\subsection{Proof of Proposition~\ref{prop:finite}}
\label{app:proof_prop2}

\paragraph{Setup.}
The $N$ candidates split into cluster $A$ (correct), of size $k$, with mean
direction $\boldsymbol\mu_A\in\Sph$, and cluster $B$ (spurious), of size $N-k$,
with mean direction $\boldsymbol\mu_B\in\Sph$; each candidate satisfies
$\E[\mathbf{u}_i]=\rho\boldsymbol\mu_A$ for $i\in A$ and $\E[\mathbf{u}_i]=\rho\boldsymbol\mu_B$
for $i\in B$, with the same concentration $\rho\in(0,1]$ as in
Assumption~\eqref{eq:assumption}. Let $\theta=\angle(\boldsymbol\mu_A,\boldsymbol\mu_B)$.

\paragraph{Step 1: the bisector direction.}
RCS assigns a candidate $\mathbf{u}$ to the cluster whose mean direction it aligns
with more, that is to $A$ iff
$\langle\mathbf{u},\boldsymbol\mu_A\rangle>\langle\mathbf{u},\boldsymbol\mu_B\rangle$,
equivalently $\langle\mathbf{u},\boldsymbol\mu_B-\boldsymbol\mu_A\rangle<0$. Define
the unit \emph{bisector direction}
\begin{align}
\mathbf{v}:=\frac{\boldsymbol\mu_B-\boldsymbol\mu_A}{\|\boldsymbol\mu_B-\boldsymbol\mu_A\|_2},
\qquad
\|\boldsymbol\mu_B-\boldsymbol\mu_A\|_2=\sqrt{2-2\cos\theta}=2\sin(\theta/2).
\end{align}
The vector $\mathbf{v}$ points from the correct cluster toward the spurious one, and
the entire cluster decision is governed by the single scalar
$\langle\mathbf{c}(P),\mathbf{v}\rangle$, with negative values favoring $A$. This is
a projection onto one fixed direction: the value of $\langle\mathbf{c}(P),\mathbf{v}\rangle$
does not depend on the ambient dimension $d$, so no dimension-dependent variance
enters the decision (in contrast to comparing the full $d$-dimensional distances,
whose orthogonal components would contribute noise growing with $d$).

\paragraph{Step 2: expected margin.}
For the geometric group with uniform weights, $\mathbf{c}(P)=\tfrac1N\sum_i\mathbf{u}_i$
and
\begin{align}
\E[\mathbf{c}(P)]
=\frac{k}{N}\,\rho\boldsymbol\mu_A+\frac{N-k}{N}\,\rho\boldsymbol\mu_B .
\end{align}
Each mean direction lies at angle $\theta/2$ from the bisector, on opposite sides, so
$\langle\boldsymbol\mu_A,\mathbf{v}\rangle=-\sin(\theta/2)$ and
$\langle\boldsymbol\mu_B,\mathbf{v}\rangle=+\sin(\theta/2)$. Hence
\begin{align}
\langle\E[\mathbf{c}(P)],\mathbf{v}\rangle
=\rho\sin(\tfrac\theta2)\Big(\tfrac{N-k}{N}-\tfrac{k}{N}\Big)
=-\rho\sin(\tfrac\theta2)\,\frac{2k-N}{N}
=-\Delta,
\qquad
\Delta:=\rho\sin(\tfrac\theta2)\frac{2k-N}{N}.
\end{align}
The expected center projects to $-\Delta$ along $\mathbf{v}$, lying on the correct
side of the bisector exactly when $\Delta>0$, that is (uniform) when $k>N/2$. This
restates the condition $k\ge N/2$ as a signed margin.

\paragraph{Step 3: concentration.}
Write the projected center as an average of scalars,
\begin{align}
\langle\mathbf{c}(P),\mathbf{v}\rangle=\frac1N\sum_{i=1}^N X_i,
\qquad X_i:=\langle\mathbf{u}_i,\mathbf{v}\rangle .
\end{align}
The $X_i$ are independent (the candidates are drawn independently), and each is
bounded, $X_i\in[-1,1]$, because $\|\mathbf{u}_i\|_2=\|\mathbf{v}\|_2=1$ and the
inner product of two unit vectors lies in $[-1,1]$. Their average has mean
$\E[\tfrac1N\sum_i X_i]=\langle\E[\mathbf{c}(P)],\mathbf{v}\rangle=-\Delta$. RCS
selects $A$ when $\tfrac1N\sum_i X_i<0$, i.e.\ when the average exceeds its mean
$-\Delta$ by less than $\Delta$. Hoeffding's inequality~\citep{hoeffding1963}, for
independent variables each bounded in an interval of length $b_i-a_i=2$, gives for
any $t>0$
\begin{align}
\Pr\!\Big(\tfrac1N\textstyle\sum_i X_i-\E\big[\tfrac1N\textstyle\sum_i X_i\big]\ge t\Big)
&\le\exp\!\Big(-\frac{2N^2t^2}{\sum_{i=1}^N(b_i-a_i)^2}\Big)\\
&=\exp\!\Big(-\frac{2N^2t^2}{4N}\Big)
=\exp\!\Big(-\frac{Nt^2}{2}\Big).
\end{align}
Setting $t=\Delta>0$,
\begin{align}
\Pr(\text{RCS selects }A)
&=\Pr\!\Big(\tfrac1N\textstyle\sum_i X_i<0\Big)\\
&=1-\Pr\!\Big(\tfrac1N\textstyle\sum_i X_i-(-\Delta)\ge\Delta\Big)\\
&\ge 1-\exp\!\big(-N\Delta^2/2\big).
\end{align}
The bound is non-asymptotic, depends only on $N$ and $\Delta$ and not on $d$, and
requires no distributional assumption beyond $\Delta>0$.

\paragraph{Step 4: other weighting variants.}

\emph{Frequency.} Replacing uniform weights by $p_i=f_i/\sum_j f_j$ changes the
deterministic part of the projected center to $\rho\sin(\tfrac\theta2)(W_B-W_A)$,
where $W_A,W_B$ are the total normalized frequency mass of clusters $A$ and $B$. The
same Hoeffding step applies with margin
$\Delta_{\mathrm{freq}}=\rho\sin(\tfrac\theta2)(W_A-W_B)$, so a lexically dominant
correct cluster ($W_A>W_B$) is recovered even when $k<N/2$.

\emph{Cosine.} Endogenous weights amplify whichever cluster is most concentrated;
when a spurious cluster is tighter than the correct one, the projected margin can
turn negative even at $k>N/2$, so no finite sample guarantee holds.

\emph{Probability.} Exogenous weights add a term proportional to the likelihood gap
between the clusters. When the likelihood favors the correct cluster, this term is
positive and adds to $\Delta$, so the correct cluster can be selected even at
$k<N/2$, the one case in which RCS recovers a genuine semantic minority. As this
depends on external likelihood calibration, we state it as a mechanism rather than a
guarantee.

\begin{remark}
When $\Delta<0$ (a semantically isolated correct minority under the geometric
group), the expected center lies on cluster $B$'s side and RCS selects the wrong
cluster, failing exactly as majority voting does. Asymptotic recovery
(Proposition~\ref{prop:asymptotic}) is preserved for all $k$ as $N\to\infty$.
\qed
\end{remark}

\subsection{Implementation Details} \label{app:details}

\paragraph{Evaluation Protocol.} For all methods, we measure accuracy (\%) between the selected answer and the ground-truth. A generation is deemed correct via exact match on long-form reasoning tasks or ROUGE-L F1 $>$ 0.3 on short-form QA tasks (SciQ, GPQA), exactly as in prior work~\citep{kuhn2023semantic, nguyen2025probabilities}. We include all model outputs, including empty or degenerate responses, to reflect real-world noisy generation settings.

We use 5-shot prompting~\citep{brown2020language} for short-form QA and Chain-of-Thought prompting~\citep{wei2022chain} for long-form tasks. For all baselines, we follow the original setup. 
We summarize the evaluation benchmarks, including the number of evaluation samples and representative examples, in Table~\ref{tab:stats}. For MMLU-Pro, we sample up to 10 questions per category (105 samples across 14 categories; some categories contain fewer than 10 eligible questions) to ensure broad coverage; for BigBenchHard, we consider two challenging tasks: Date Understanding (\textbf{BBH-Date}) and Navigate (\textbf{BBH-Navigate}). 
We observe no differences between breaking ties randomly and preserving the original answer order. For determinism and reproducibility, we adopt the latter. 
For the multi-agent debate setting, we follow prior implementations~\citep{choi2025debate, nguyen2026hear}. Sampling hyperparameters and generation prompts are provided in Tables~\ref{tab:params} and~\ref{tab:prompt}, respectively. All experiments are conducted with three different seeds (42, 44, and 46) on a single H100-80GB GPU. Total compute across all models and benchmarks is approximately 120 GPU-hours; a single model evaluation (one benchmark, one model, $N{=}10$) completes in roughly 2--5 hours depending on model size.


\begin{table}[ht]
\centering
\caption{Sampling hyperparameters used for generation.}
\label{tab:params}
\begin{tabular}{l | c}
\toprule
\textbf{Parameter} & \textbf{Value} \\
\midrule
Temperature & 1 \\
Max new tokens (short-form QA) & 32 \\
Max new tokens (long-form tasks) & 512 \\
\bottomrule
\end{tabular}
\end{table}

\begin{table}[ht]
\centering
\caption{System prompts used for generation across tasks.}
\label{tab:prompt}

\begin{tabular}{p{0.33\linewidth} p{0.59\linewidth}}
\toprule
\textbf{Tasks} & \textbf{Prompt} \\
\midrule

SciQ, GPQA 
& This is a bot that correctly answers questions. \\

\midrule
Arithmetics, GSM8K 
& Make sure to state your final answer in curly brackets at the very end of your response, just like:
\texttt{\{final answer: 12.34\}}. Let's think step by step. \\

\midrule
Form.Log., MMLU-Pro, BBH-Date 
& Make sure to state your final answer choice in curly brackets at the very end of your response, just like:
\texttt{\{final answer: (A)\}}. Let's think step by step. \\

\midrule
BBH-Navigate
& Make sure to state your final answer choice in curly brackets at the very end of your response, just like:
\texttt{\{final answer: Yes\}}. Let's think step by step. \\

\bottomrule
\end{tabular}

\end{table}

\begin{table*}[ht]
\centering
\caption{Overview of evaluation benchmarks, including the number of samples and representative examples.}
\label{tab:stats}
\resizebox{\textwidth}{!}{
\begin{tabular}{l | c |p{6cm} |p{4cm}}
\toprule
\textbf{Dataset} & \textbf{\#Samples} & \textbf{Question} & \textbf{Answer} \\
\midrule

SciQ & 1000 & 
Compounds that are capable of accepting electrons, such as o 2 or f2, are called what?
& oxidants \\
\midrule

GPQA & 198 &
Which of the following physical theories never requires UV regularization?
& Superstring Theory \\
\midrule

Arithmetics & 200 & 
What is $23 + 47$? 
& 70 \\
\midrule

GSM8K & 200 & 
A robe takes 2 bolts of blue fiber and half that much white fiber. How many bolts in total does it take? 
& 3 \\
\midrule


Formal Logic & 126 & 
Select the best translation into predicate logic: Sheena is a punk rocker. 
\newline \textit{Choices:} ("Sx", "xS", "sP", "Ps") 
& "Ps" \\
\midrule

MMLU-Pro & 105 & 
Determine the number of men needed to build a boat in 77 days if it takes 36 men 132 days to build one.
\newline \textit{Choices:} ("84 men",
"36 men",
"99 men",
"132 men",
"45 men",
"70 men",
"62 men",
"50 men",
"77 men",
"120 men"
)
& "62 men" \\
\midrule

BBH-Date & 250 & 
Today is Christmas Eve of 1937. What is the date tomorrow in MM/DD/YYYY?
\newline \textit{Options:}
(A) 12/11/1937
(B) 12/25/1937
(C) 01/04/1938
(D) 12/04/1937
(E) 12/25/2006
(F) 07/25/1937
&  (B)\\
\midrule

BBH-Navigate & 250 & 
If you follow these instructions, do you return to the starting point? Always face forward. Take 1 step backward. Take 9 steps left. Take 2 steps backward. Take 6 steps forward. Take 4 steps forward. Take 4 steps backward. Take 3 steps right.
\newline \textit{Options:}
- Yes
- No
&  No\\



\bottomrule
\end{tabular}}
\end{table*}


\begin{table}[t]
  \centering
  \caption{Extended Results: Best-of-$N$ selection accuracy across settings. Best-of-$N$ selection accuracy. \textbf{BBF}: black-box friendly. For non-RCS methods, \underline{underline} marks the best baseline per setting. Among RCS variants, \textbf{bold} indicates the best RCS method per column. No baseline significantly outperforms RCS at the column level ($p<0.05$, paired t-test); a full per-cell (model $\times$ benchmark) McNemar breakdown against all six baselines is in Appendix Table~\ref{tab:r9-full-per-cell}.
}
  \label{tab:ranking-addition}
  \resizebox{\columnwidth}{!}{
  \begin{tabular}{ll |c| ccccc | c}
    \toprule
    \textbf{Model} & \textbf{Method} & \textbf{BBF}
    & \textbf{SciQ} & \textbf{GPQA} 
    & \textbf{Arithmetics} & \textbf{GSM8K} & \textbf{Form.Log.} 
    & \textbf{Average}
    \\
    \midrule

    \multirow{13}{*}{Qwen2.5-7B}
    & Greedy & \ding{51}& 64.1 & 20.5 & 97.0 & 88.3 & 43.6 & 63.1\\
    & Oracle & \ding{51} & 	82.0$\pm$1.1 &	49.9$\pm$1.6 & 	100.0$\pm$0.0	& 94.6$\pm$0.8	& 67.7$\pm$2.5 & 78.8 \\
    \cmidrule(lr){2-9}
    & NLL & \ding{55}& 66.4$\pm$0.1 & 18.3$\pm$1.3 & 47.3$\pm$5.7 & 32.7$\pm$4.3 & 25.7$\pm$0.5 & 38.1 \\
    & ANLL & \ding{55}& 66.2$\pm$0.4 & 18.1$\pm$0.9 & 47.0$\pm$5.6 & 32.7$\pm$4.4 & 25.9$\pm$0.5 & 38.0 \\
    & SC & \ding{51}   & 70.3$\pm$0.9 & 22.1$\pm$4.2 & 74.3$\pm$2.0 & \underline{52.3$\pm$2.0} & 48.4$\pm$0.0 & 53.5 \\
    & SCW & \ding{51} & 70.2$\pm$0.8 & \underline{26.3$\pm$3.2} & 52.8$\pm$5.0&	36.4$\pm$5.2	&32.8$\pm$3.3 & 43.7\\
    & CE & \ding{55}   & \underline{70.4$\pm$0.4} & 21.6$\pm$3.8 & \underline{75.5$\pm$4.0} & 51.9$\pm$3.1 & \underline{48.7$\pm$1.2} & 53.6 \\
    & ModeX & \ding{51} & 69.0$\pm$0.6	&21.6$\pm$1.6	&54.3$\pm$5.1	&37.6$\pm$3.3&	35.2$\pm$2.6& 43.5\\
    &\cellcolor{green!15} RCS$_\text{medoid}$ &\cellcolor{green!15} \ding{51} &\cellcolor{green!15} 70.1$\pm$0.6 &\cellcolor{green!15} 25.0$\pm$1.3 &\cellcolor{green!15} 77.5$\pm$3.6 &\cellcolor{green!15} 57.5$\pm$2.6 &\cellcolor{green!15} \textbf{49.2$\pm$2.9} &\cellcolor{green!15} 55.9 \\
    &\cellcolor{green!15} RCS$_\text{uni}$ &\cellcolor{green!15} \ding{51} & \cellcolor{green!15} 70.2$\pm$0.2 & \cellcolor{green!15}\textbf{27.1$\pm$1.7} &\cellcolor{green!15} 77.5$\pm$2.3 &\cellcolor{green!15} 61.6$\pm$2.8 &\cellcolor{green!15} \textbf{49.2$\pm$3.2} &\cellcolor{green!15} 57.1 \\
    & \cellcolor{green!15} RCS$_\text{freq}$ &\cellcolor{green!15} \ding{51} &\cellcolor{green!15} \textbf{70.3$\pm$0.5} &\cellcolor{green!15} 25.9$\pm$1.8 &\cellcolor{green!15} 76.7$\pm$1.4 &\cellcolor{green!15} 56.0$\pm$2.1 &\cellcolor{green!15} 47.9$\pm$2.4 &\cellcolor{green!15} 55.4 \\
    & \cellcolor{green!15} RCS$_\text{cosine}$ &\cellcolor{green!15} \ding{51} &\cellcolor{green!15} 70.1$\pm$0.5 &\cellcolor{green!15} 26.9$\pm$1.8 &\cellcolor{green!15} 76.8$\pm$3.4 &\cellcolor{green!15} 61.3$\pm$3.1 &\cellcolor{green!15} \textbf{49.2$\pm$3.2} &\cellcolor{green!15} 56.9\\
    & \cellcolor{green!15} RCS$_\text{prob}$ &\cellcolor{green!15} \ding{55} &\cellcolor{green!15} 70.0$\pm$0.4 &\cellcolor{green!15} \textbf{27.1$\pm$2.4} &\cellcolor{green!15} \textbf{77.7$\pm$1.9} &\cellcolor{green!15} \textbf{62.1$\pm$3.1} &\cellcolor{green!15} \textbf{49.2$\pm$3.2} & \cellcolor{green!15}\textbf{57.2} \\
    \midrule

    \multirow{13}{*}{Llama3.1-8B}
    & Greedy & \ding{51}& 63.1 & 25.5 & 89.5 & 88.3 & 34.9 & 60.8\\
    & Oracle & \ding{51} & 85.1$\pm$0.4	& 56.3$\pm$1.3 & 99.3$\pm$0.3	& 94.6$\pm$0.8	& 74.1$\pm$1.2 & 81.9 \\
    \cmidrule(lr){2-9}
    & NLL & \ding{55}& 61.0$\pm$0.5 & 20.4$\pm$0.8 & 73.3$\pm$2.8 & \underline{69.0$\pm$0.5} & \underline{33.9$\pm$2.8} & 51.5 \\
    & ANLL & \ding{55}& 61.1$\pm$0.5 & 18.7$\pm$0.9 & 73.5$\pm$4.1 & 68.7$\pm$1.6 & 33.1$\pm$2.4 & 51.0 \\
    & SC & \ding{51}   & 67.3$\pm$0.7 & 16.8$\pm$1.8 & 78.3$\pm$3.0 & 66.0$\pm$2.3 & 31.7$\pm$2.1 & 52.0 \\
    & SCW & \ding{51} & \underline{67.7$\pm$1.0} & \underline{26.8$\pm$0.8} & 53.8$\pm$4.2&	50.3$\pm$5.9&	25.7$\pm$1.8 & 44.9 \\
    & CE & \ding{55}   & 66.8$\pm$0.9 & 18.5$\pm$1.1 & \underline{81.0$\pm$3.8} & 67.5$\pm$1.3 & 32.3$\pm$1.7 & 53.2 \\
    & ModeX & \ding{51} &	63.1$\pm$1.2&	17.8$\pm$1.1&	55.3$\pm$2.5	&48.4$\pm$3.9	&25.1$\pm$2.6 & 41.9\\
    &\cellcolor{green!15} RCS$_\text{medoid}$ &\cellcolor{green!15} \ding{51} &\cellcolor{green!15} 68.0$\pm$0.9 &\cellcolor{green!15} 26.4$\pm$0.9 &\cellcolor{green!15} 86.5$\pm$1.5 &\cellcolor{green!15} 71.4$\pm$2.8 &\cellcolor{green!15} 37.8$\pm$0.9 &\cellcolor{green!15} 58.0 \\
    &\cellcolor{green!15} RCS$_\text{uni}$ &\cellcolor{green!15} \ding{51} &\cellcolor{green!15} \textbf{68.1$\pm$0.7} &\cellcolor{green!15} 26.9$\pm$1.0 &\cellcolor{green!15} \textbf{88.5$\pm$0.5} &\cellcolor{green!15} 72.8$\pm$3.0 &\cellcolor{green!15} \textbf{39.4$\pm$0.9} &\cellcolor{green!15} \textbf{59.1} \\
    & \cellcolor{green!15} RCS$_\text{freq}$ &\cellcolor{green!15} \ding{51} &\cellcolor{green!15} 67.6$\pm$1.1 &\cellcolor{green!15} 26.9$\pm$1.0 &\cellcolor{green!15} 83.7$\pm$2.3 & \cellcolor{green!15} 68.9$\pm$1.9 & \cellcolor{green!15} 34.9$\pm$0.8& \cellcolor{green!15} 56.4 \\
    & \cellcolor{green!15} RCS$_\text{cosine}$ &\cellcolor{green!15} \ding{51} &\cellcolor{green!15} \textbf{68.1$\pm$0.7} &\cellcolor{green!15} 27.3$\pm$0.8 &\cellcolor{green!15} 87.7$\pm$1.2 &\cellcolor{green!15} 72.8$\pm$2.3 &\cellcolor{green!15} \textbf{39.4$\pm$0.9} &\cellcolor{green!15} \textbf{59.1} \\
    & \cellcolor{green!15} RCS$_\text{prob}$ &\cellcolor{green!15} \ding{55} &\cellcolor{green!15} 67.9$\pm$1.1 &\cellcolor{green!15} \textbf{27.5$\pm$0.9} &\cellcolor{green!15} 85.2$\pm$0.6 &\cellcolor{green!15} \textbf{75.6$\pm$1.0} &\cellcolor{green!15} 36.8$\pm$2.6 &\cellcolor{green!15} 58.6 \\
    \midrule

    \multirow{13}{*}{Gemma2-9B}
    & Greedy & \ding{51}& 67.6 & 22.6 & 96.0 & 69.0 & 52.4 & 62.0\\
    & Oracle & \ding{51} & 84.6$\pm$0.6 &	48.5$\pm$4.2 & 98.2$\pm$0.6	& 95.3$\pm$0.3	& 77.8$\pm$5.7 & 80.9\\
    \cmidrule(lr){2-9}
    & NLL & \ding{55} & 72.9$\pm$1.1 & \underline{25.6$\pm$1.1} & 94.8$\pm$1.3 & 67.9$\pm$1.1 & 48.1$\pm$2.8 & 61.9 \\
    & ANLL & \ding{55}& 66.3$\pm$0.9 & 21.2$\pm$2.3 & 96.0$\pm$1.0 & \underline{73.1$\pm$0.9} & 50.0$\pm$0.8 & 61.3 \\
    & SC & \ding{51}   & 73.9$\pm$0.2 & 19.7$\pm$0.9 & 97.3$\pm$0.3 & 72.9$\pm$3.5 & 51.6$\pm$0.8 & 63.1 \\
    & SCW & \ding{51} & \underline{74.0$\pm$0.6} & 25.4$\pm$0.5 & 92.2$\pm$1.0	&60.1$\pm$1.8	&48.9$\pm$3.2 & 60.1\\
    & CE & \ding{55}   & 73.6$\pm$0.6 & 18.6$\pm$0.5 & \underline{97.5$\pm$0.0} & \underline{73.1$\pm$3.7} & \underline{52.1$\pm$1.7} & 63.0 \\
    & ModeX & \ding{51} & 71.7$\pm$0.2	&19.5$\pm$2.2	&92.7$\pm$2.8	&62.6$\pm$1.5&	49.2$\pm$4.2& 59.1\\
    &\cellcolor{green!15} RCS$_\text{medoid}$ &\cellcolor{green!15} \ding{51} &\cellcolor{green!15} \textbf{74.2$\pm$0.4} &\cellcolor{green!15} 24.4$\pm$1.4 &\cellcolor{green!15} 97.2$\pm$0.6 &\cellcolor{green!15} 72.9$\pm$3.5 &\cellcolor{green!15} 52.6$\pm$0.9 &\cellcolor{green!15} 64.3 \\
    & \cellcolor{green!15} RCS$_\text{uni}$ &\cellcolor{green!15} \ding{51} &\cellcolor{green!15} 74.0$\pm$0.4 & \cellcolor{green!15}25.9$\pm$0.5 &\cellcolor{green!15}97.2$\pm$0.6 &\cellcolor{green!15} 73.8$\pm$2.8 &\cellcolor{green!15} \textbf{52.9$\pm$2.0} &\cellcolor{green!15} 64.8 \\ 
    & \cellcolor{green!15} RCS$_\text{freq}$ &\cellcolor{green!15} \ding{51} &\cellcolor{green!15} 73.9$\pm$0.5 &\cellcolor{green!15} 25.7$\pm$0.6 &\cellcolor{green!15} \textbf{97.3$\pm$0.3} &\cellcolor{green!15} 72.6$\pm$3.7 &\cellcolor{green!15} 51.8$\pm$1.2 &\cellcolor{green!15} 64.3 \\
    & \cellcolor{green!15} RCS$_\text{cosine}$ &\cellcolor{green!15} \ding{51} &\cellcolor{green!15} 74.1$\pm$0.4 &\cellcolor{green!15} 25.7$\pm$0.3 &\cellcolor{green!15} 97.2$\pm$0.6 &\cellcolor{green!15} 71.1$\pm$4.5 &\cellcolor{green!15} 51.8$\pm$1.2 &\cellcolor{green!15} 64.0\\
    & \cellcolor{green!15} RCS$_\text{prob}$ &\cellcolor{green!15} \ding{55} &\cellcolor{green!15} \textbf{74.2$\pm$0.5} &\cellcolor{green!15} \textbf{26.3$\pm$0.6} &\cellcolor{green!15} \textbf{97.3$\pm$0.3} &\cellcolor{green!15} \textbf{75.1$\pm$3.5} &\cellcolor{green!15} 52.4$\pm$0.8 &\cellcolor{green!15} \textbf{65.1} \\
    \bottomrule
  \end{tabular}
  }
\end{table}

\begin{table*}[t]
\centering
\caption{Full per-cell (model $\times$ benchmark) item-level McNemar $p$-value of RCS$_\text{uni}$ against each of the six baselines. \textbf{Bold} = significant at $\alpha{=}0.05$ in RCS$_\text{uni}$'s favor; \textcolor{red}{\textbf{red bold}} = significant in the baseline's favor.}
\label{tab:r9-full-per-cell}
\resizebox{\textwidth}{!}{
\begin{tabular}{llccccc}
\toprule
Model & Baseline & SciQ & GPQA & Arithmetics & GSM8K & Form.Log. \\
\midrule
\multirow{6}{*}{Qwen2.5-3B} & NLL & \textbf{\boldmath $<10^{-20}$} & \textbf{\boldmath $<10^{-4}$} & \textbf{\boldmath $<10^{-74}$} & \textbf{\boldmath $<10^{-54}$} & \textbf{\boldmath $<10^{-16}$} \\
 & ANLL & \textbf{\boldmath $<10^{-21}$} & \textbf{\boldmath $<10^{-4}$} & \textbf{\boldmath $<10^{-74}$} & \textbf{\boldmath $<10^{-54}$} & \textbf{\boldmath $<10^{-17}$} \\
 & SC & \textbf{\boldmath 0.001} & 0.326 & \textbf{\boldmath $<10^{-7}$} & \textbf{\boldmath $<10^{-18}$} & \textbf{\boldmath 0.003} \\
 & SCW & \textbf{\boldmath 0.001} & 1.000 & 0.108 & \textbf{\boldmath $<10^{-6}$} & 0.388 \\
 & CE & \textbf{\boldmath $<10^{-4}$} & \textbf{\boldmath $<10^{-4}$} & \textbf{\boldmath $<10^{-4}$} & \textbf{\boldmath $<10^{-19}$} & \textbf{\boldmath 0.036} \\
 & ModeX & \textbf{\boldmath $<10^{-9}$} & \textbf{\boldmath 0.007} & \textbf{\boldmath $<10^{-39}$} & \textbf{\boldmath $<10^{-28}$} & \textbf{\boldmath $<10^{-6}$} \\
\midrule
\multirow{6}{*}{Qwen2.5-7B} & NLL & \textbf{\boldmath $<10^{-11}$} & \textbf{\boldmath $<10^{-6}$} & \textbf{\boldmath $<10^{-29}$} & \textbf{\boldmath $<10^{-27}$} & \textbf{\boldmath $<10^{-18}$} \\
 & ANLL & \textbf{\boldmath $<10^{-13}$} & \textbf{\boldmath $<10^{-6}$} & \textbf{\boldmath $<10^{-29}$} & \textbf{\boldmath $<10^{-27}$} & \textbf{\boldmath $<10^{-18}$} \\
 & SC & 0.901 & \textbf{\boldmath 0.002} & \textbf{\boldmath $<10^{-4}$} & \textbf{\boldmath $<10^{-13}$} & 0.815 \\
 & SCW & 0.766 & 0.839 & 0.481 & \textbf{\boldmath $<10^{-4}$} & 1.000 \\
 & CE & 0.665 & \textbf{\boldmath 0.001} & \textbf{\boldmath 0.029} & \textbf{\boldmath $<10^{-11}$} & 1.000 \\
 & ModeX & \textbf{\boldmath 0.011} & \textbf{\boldmath 0.003} & \textbf{\boldmath $<10^{-15}$} & \textbf{\boldmath $<10^{-17}$} & \textbf{\boldmath $<10^{-10}$} \\
\midrule
\multirow{6}{*}{Llama3.2-3B} & NLL & \textbf{\boldmath $<10^{-16}$} & \textbf{\boldmath 0.009} & \textbf{\boldmath $<10^{-16}$} & \textbf{\boldmath 0.000} & \textcolor{red}{\textbf{\boldmath $<10^{-6}$}} \\
 & ANLL & \textbf{\boldmath $<10^{-16}$} & \textbf{\boldmath 0.004} & \textbf{\boldmath $<10^{-12}$} & \textbf{\boldmath $<10^{-4}$} & \textcolor{red}{\textbf{\boldmath 0.000}} \\
 & SC & 0.080 & \textbf{\boldmath $<10^{-5}$} & 0.344 & \textbf{\boldmath 0.002} & \textbf{\boldmath 0.000} \\
 & SCW & 1.000 & 0.791 & 1.000 & 0.096 & \textbf{\boldmath 0.002} \\
 & CE & 0.750 & \textbf{\boldmath 0.022} & 0.227 & \textbf{\boldmath 0.011} & \textbf{\boldmath 0.012} \\
 & ModeX & \textbf{\boldmath $<10^{-16}$} & \textbf{\boldmath $<10^{-4}$} & \textbf{\boldmath $<10^{-40}$} & \textbf{\boldmath $<10^{-19}$} & 0.091 \\
\midrule
\multirow{6}{*}{Llama3.1-8B} & NLL & \textbf{\boldmath $<10^{-24}$} & \textbf{\boldmath 0.000} & \textbf{\boldmath $<10^{-12}$} & 0.078 & \textbf{\boldmath 0.018} \\
 & ANLL & \textbf{\boldmath $<10^{-23}$} & \textbf{\boldmath $<10^{-5}$} & \textbf{\boldmath $<10^{-12}$} & 0.053 & \textbf{\boldmath 0.008} \\
 & SC & 0.056 & \textbf{\boldmath $<10^{-7}$} & \textbf{\boldmath $<10^{-15}$} & \textbf{\boldmath $<10^{-10}$} & \textbf{\boldmath $<10^{-6}$} \\
 & SCW & 0.201 & 1.000 & \textbf{\boldmath 0.000} & \textbf{\boldmath 0.001} & \textbf{\boldmath 0.003} \\
 & CE & \textbf{\boldmath 0.003} & \textbf{\boldmath $<10^{-6}$} & \textbf{\boldmath $<10^{-8}$} & \textbf{\boldmath $<10^{-5}$} & \textbf{\boldmath $<10^{-5}$} \\
 & ModeX & \textbf{\boldmath $<10^{-21}$} & \textbf{\boldmath $<10^{-4}$} & \textbf{\boldmath $<10^{-44}$} & \textbf{\boldmath $<10^{-21}$} & \textbf{\boldmath $<10^{-7}$} \\
\midrule
\multirow{6}{*}{Gemma2-9B} & NLL & \textbf{\boldmath 0.016} & 0.902 & \textbf{\boldmath 0.001} & \textbf{\boldmath 0.006} & \textbf{\boldmath 0.012} \\
 & ANLL & \textbf{\boldmath $<10^{-37}$} & \textbf{\boldmath 0.003} & 0.065 & 1.000 & 0.132 \\
 & SC & 0.834 & \textbf{\boldmath $<10^{-4}$} & 1.000 & 0.500 & 0.125 \\
 & SCW & 1.000 & 0.581 & 1.000 & \textcolor{red}{\textbf{\boldmath 0.007}} & 0.125 \\
 & CE & 0.337 & \textbf{\boldmath $<10^{-5}$} & 0.500 & 1.000 & 0.625 \\
 & ModeX & \textbf{\boldmath $<10^{-6}$} & \textbf{\boldmath $<10^{-4}$} & \textbf{\boldmath $<10^{-6}$} & \textbf{\boldmath $<10^{-5}$} & 0.542 \\
\bottomrule
\end{tabular}
}
\end{table*}


\begin{figure*}[ht]
    \centering
    \includegraphics[width=\linewidth]{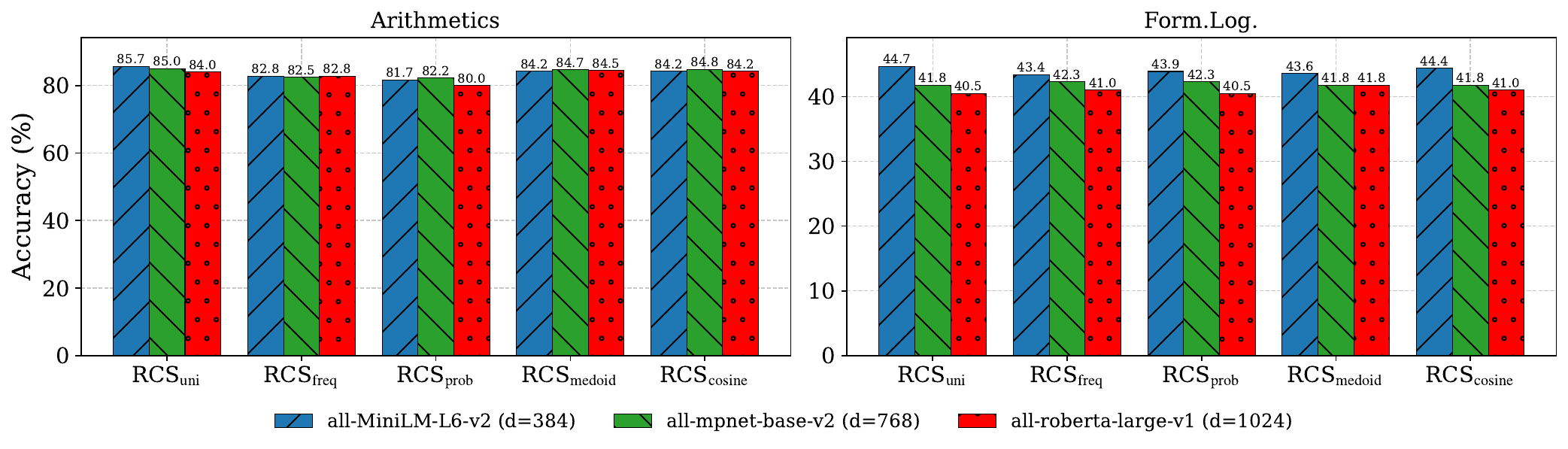}
    \caption{
    Effect of the sentence embedding model on Arithmetics and Form.Log. using Qwen2.5-3B.
    }
    \label{fig:embed_qwen2.5-3b}
\end{figure*}

\begin{figure*}[ht]
    \centering
    \includegraphics[width=\linewidth]{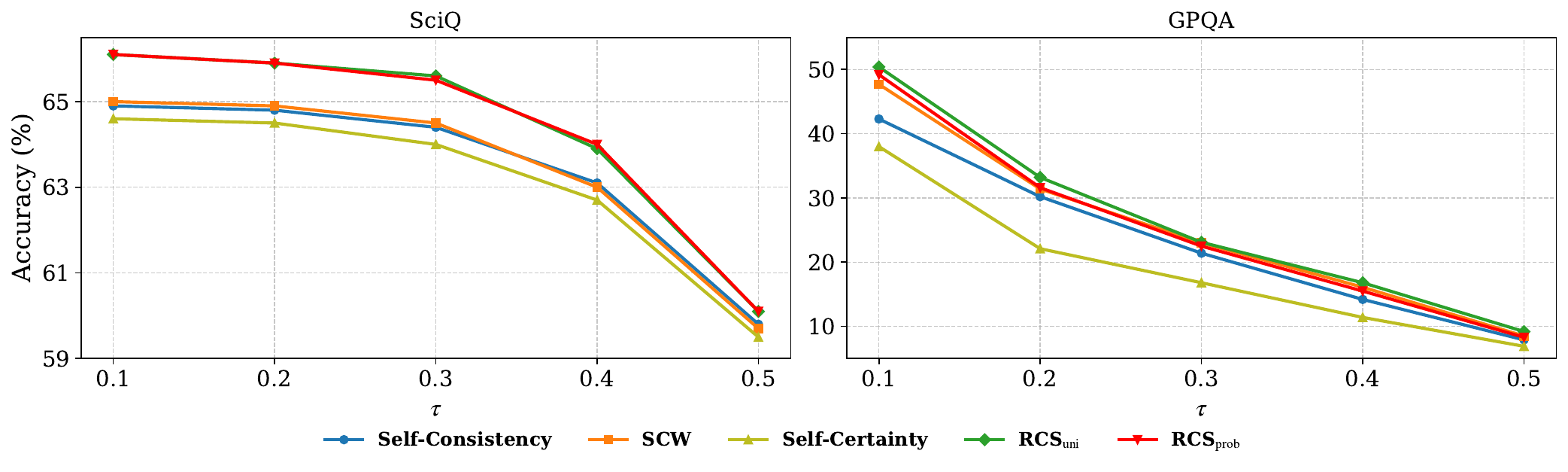}
    \caption{
    Performance when varying correctness threshold ($\tau$) using Llama3.2-3B.
    }
    \label{fig:threshold_llama3.2-3b}
\end{figure*}




\subsection{Extended Results: Number Of Samples}\label{app:sample-details}

Tables~\ref{tab:ranking_n5},~\ref{tab:ranking_n20}, and~\ref{tab:ranking_n40} report performance over different number of sampling responses. 

\begin{table}[t]
  \centering
  \caption{Best-of-$N$ selection accuracy across settings ($N{=}5$).}
  \label{tab:ranking_n5}
  \begin{tabular}{ll|ccccc}
    \toprule
    \textbf{Model} & \textbf{Method}
    & \textbf{SciQ} & \textbf{GPQA} 
    & \textbf{Arithmetics} & \textbf{GSM8K} & \textbf{Form.Log.} 
    \\
    \midrule
    \multirow{11}{*}{Qwen2.5-3B}
    & NLL & 61.7$\pm$1.2 &	17.6$\pm$3.1	& 49.2$\pm$12.4	& 45.2$\pm$4.7 &	28.0$\pm$3.2\\
    & ANLL & 61.6$\pm$1.1	& 17.1$\pm$4.1	& 48.7$\pm$11.4	& 46.0$\pm$4.8	& 27.5$\pm$3.3\\
    & SC  & 63.8$\pm$0.5	& 18.1$\pm$2.7	& 78.7$\pm$7.2	& 65.0$\pm$0.5	& 39.4$\pm$1.8\\
    & SCW & 64.0$\pm$0.2 & 24.0$\pm$2.6 & 46.8$\pm$1.1	&38.1$\pm$6.2&	27.0$\pm$2.1\\
    & CE    & 63.6$\pm$0.3	& 15.7$\pm$1.7	& 78.5$\pm$7.4	& 66.8$\pm$5.0	& 42.4$\pm$0.2\\
    & ModeX    & 61.2$\pm$1.2	& 18.0$\pm$2.4	& 46.0$\pm$3.0	& 32.0$\pm$0.5	& 24.9$\pm$0.9\\
    & RCS$_\text{medoid}$  &64.0$\pm$0.4	& 24.7$\pm$2.2	& 77.7$\pm$2.8	& 66.7$\pm$2.9	& 40.7$\pm$0.5\\
    & RCS$_\text{uni}$& 63.8$\pm$0.1	& 24.2$\pm$2.9	& 77.0$\pm$2.8	& 66.5$\pm$4.0	& 40.7$\pm$0.5 \\
    &  RCS$_\text{freq}$&64.0$\pm$0.2	& 24.0$\pm$2.6	& 77.5$\pm$3.1	& 66.3$\pm$3.8	& 41.0$\pm$0.5\\
    & RCS$_\text{cosine}$ & 63.9$\pm$0.2 & 24.9$\pm$2.4 & 72.8$\pm$2.5 & 55.5$\pm$2.4 & 36.5$\pm$1.4\\
    &  RCS$_\text{prob}$  &64.0$\pm$0.5	& 21.6$\pm$2.0	& 75.7$\pm$5.3	& 63.1$\pm$4.4	& 41.3$\pm$0.0\\
    \midrule
    
    \multirow{11}{*}{Qwen2.5-7B}
    & NLL & 67.6$\pm$1.5	& 17.1$\pm$1.4	& 63.3$\pm$10.9	& 51.4$\pm$3.4	& 36.0$\pm$2.3\\
    & ANLL & 67.5$\pm$1.1	& 16.9$\pm$1.2	& 63.5$\pm$11.0	& 51.4$\pm$3.4	& 35.4$\pm$2.0\\
    & SC  & 69.2$\pm$0.5	& 22.3$\pm$3.2	& 83.3$\pm$13.7	& 72.4$\pm$3.1	& 47.4$\pm$4.4\\
    & SCW & 68.9$\pm$0.4 & 26.9$\pm$0.9 & 53.2$\pm$1.1	&38.9$\pm$2.8	&32.0$\pm$1.2\\
    & CE    & 69.4$\pm$0.1	& 19.9$\pm$1.1	& 84.0$\pm$16.0	& 74.3$\pm$2.6	& 47.8$\pm$3.6\\
    & ModeX & 69.0$\pm$0.7 & 22.1$\pm$3.0 & 54.2$\pm$3.2 & 40.6$\pm$2.7 & 34.1$\pm$6.3\\
    & RCS$_\text{medoid}$  &69.0$\pm$0.4	& 27.3$\pm$0.8	&84.3$\pm$12.0	&74.3$\pm$1.8	&47.6$\pm$3.6\\
    & RCS$_\text{uni}$&68.9$\pm$0.4	&27.6$\pm$1.3	&84.2$\pm$11.8	&75.0$\pm$1.3	&47.6$\pm$3.6\\
    &  RCS$_\text{freq}$&69.0$\pm$0.4	&26.9$\pm$0.9	&84.3$\pm$12.0	&75.1$\pm$1.0	&47.6$\pm$3.6\\
    & RCS$_\text{cosine}$ & 69.0$\pm$0.4 & 27.5$\pm$1.0 & 74.8$\pm$6.0 & 60.1$\pm$2.4 & 45.2$\pm$4.4\\
    &  RCS$_\text{prob}$  &69.3$\pm$0.4	&25.2$\pm$1.5	&83.0$\pm$11.8	&70.9$\pm$2.0	&47.9$\pm$3.3\\
    \midrule

    \multirow{11}{*}{Llama3.2-3B}
    & NLL & 54.0$\pm$1.3	&19.2$\pm$0.9	&82.7$\pm$4.2	&73.4$\pm$3.8	&31.0$\pm$2.9\\
    & ANLL & 53.7$\pm$0.6	&18.5$\pm$2.0	&86.7$\pm$3.3	&76.1$\pm$1.8	&31.5$\pm$2.0\\
    & SC  & 58.6$\pm$0.8	&16.1$\pm$2.4	&93.3$\pm$2.1	&79.2$\pm$3.6	&31.5$\pm$1.7\\
    & SCW & 59.4$\pm$1.2 & 22.3$\pm$2.6 & 69.8$\pm$0.4&	59.7$\pm$1.3&	16.9$\pm$3.9\\
    & CE    & 57.6$\pm$1.4	&19.2$\pm$3.2	&93.0$\pm$1.8	&80.3$\pm$4.1	&33.7$\pm$2.4\\
    & ModeX    & 53.4$\pm$1.1&	12.8$\pm$3.2	&68.8$\pm$1.0	&61.3$\pm$2.6	&19.1$\pm$3.5\\
    & RCS$_\text{medoid}$  &59.4$\pm$1.2	&22.5$\pm$2.9	&94.0$\pm$0.5	&80.2$\pm$2.0	&30.9$\pm$4.8\\
    & RCS$_\text{uni}$& 59.3$\pm$1.5	&22.3$\pm$2.6	&93.3$\pm$0.6	&80.4$\pm$2.1	&30.9$\pm$4.8\\
    &  RCS$_\text{freq}$&59.4$\pm$1.2	&22.3$\pm$2.6	&93.8$\pm$0.8	&80.4$\pm$2.1	&30.9$\pm$4.8\\
    & RCS$_\text{cosine}$ & 59.3$\pm$1.4 & 22.5$\pm$2.9 & 92.8$\pm$1.1 & 76.3$\pm$1.6 & 20.9$\pm$4.4\\
    &  RCS$_\text{prob}$  &58.9$\pm$1.4	&21.2$\pm$3.6	&93.5$\pm$1.5	&80.0$\pm$1.9	&32.8$\pm$2.4\\
    \midrule

    \multirow{11}{*}{Llama3.1-8B}
    & NLL & 61.2$\pm$0.6	&21.1$\pm$1.1&	73.3$\pm$2.5&	72.8$\pm$3.3&	38.9$\pm$1.4\\
    & ANLL & 60.8$\pm$0.5	&20.7$\pm$1.4	&73.2$\pm$3.3	&73.3$\pm$3.1&	38.4$\pm$1.8\\
    & SC  & 66.0$\pm$1.8	&19.2$\pm$1.9&	82.2$\pm$7.2&	82.6$\pm$3.8	&43.1$\pm$3.9\\
    & SCW & 67.1$\pm$0.7 & 26.8$\pm$1.3 & 52.8$\pm$3.3&	55.7$\pm$3.1	&28.6$\pm$3.6\\
    & CE    & 65.6$\pm$0.4	&19.7$\pm$3.2	&82.1$\pm$8.4	&82.4$\pm$3.1	&42.4$\pm$3.0\\
    & ModeX    & 62.1$\pm$0.9	&20.5$\pm$2.1	&50.2$\pm$0.8	&53.5$\pm$3.8	&25.1$\pm$0.5\\
    & RCS$_\text{medoid}$  &67.0$\pm$0.8	&27.3$\pm$1.3	&86.0$\pm$4.0	&83.2$\pm$5.2	&41.8$\pm$3.2\\
    & RCS$_\text{uni}$& 67.0$\pm$0.6	&26.9$\pm$2.1	&85.5$\pm$3.0	&83.2$\pm$5.6	&42.3$\pm$3.6\\
    &  RCS$_\text{freq}$&67.1$\pm$0.7	&26.8$\pm$1.3	&85.7$\pm$3.6	&83.2$\pm$5.6	&42.3$\pm$3.6\\
    & RCS$_\text{cosine}$ & 66.9$\pm$0.6 & 27.1$\pm$2.1 & 81.3$\pm$0.8 & 70.9$\pm$2.4 & 36.8$\pm$2.6\\
    &  RCS$_\text{prob}$  &66.7$\pm$0.8	&26.4$\pm$2.7	&77.3$\pm$3.6	&76.0$\pm$3.1	&39.2$\pm$4.1\\
    \midrule

    \multirow{11}{*}{Gemma2-9B}
    & NLL & 69.2$\pm$0.4	&27.1$\pm$1.1	&94.5$\pm$1.0	&87.3$\pm$0.9	&54.5$\pm$4.4\\
    & ANLL & 66.2$\pm$0.9	&21.1$\pm$1.1	&95.8$\pm$1.0	&87.8$\pm$1.3	&54.8$\pm$2.1\\
    & SC  & 72.8$\pm$0.6	&17.3$\pm$2.3	&97.0$\pm$0.9	&88.8$\pm$1.8	&55.6$\pm$5.0\\
    & SCW & 73.9$\pm$0.6 & 24.9$\pm$2.3 & 92.0$\pm$1.8&	65.7$\pm$2.5	&47.6$\pm$5.6\\

    & CE    & 73.2$\pm$0.4	&17.8$\pm$3.5	&97.0$\pm$0.9&	89.8$\pm$2.1	&55.5$\pm$5.2\\
    & ModeX    & 71.5$\pm$0.6	&20.7$\pm$0.5	&92.5$\pm$1.0	&64.0$\pm$2.3	&52.1$\pm$6.5\\

    & RCS$_\text{medoid}$  &73.7$\pm$0.7&	24.2$\pm$2.0&	96.8$\pm$0.6	&89.0$\pm$2.1&	55.3$\pm$4.8\\
    & RCS$_\text{uni}$& 73.8$\pm$0.4	&25.9$\pm$2.1	&96.8$\pm$0.6	&88.7$\pm$1.6	&55.8$\pm$4.8\\
    &  RCS$_\text{freq}$&74.0$\pm$0.6	&24.9$\pm$2.3	&96.8$\pm$0.6	&88.7$\pm$1.6	&55.6$\pm$4.4\\
    & RCS$_\text{cosine}$ & 73.8$\pm$0.5 & 25.7$\pm$2.2 & 96.8$\pm$0.6 & 72.8$\pm$1.1 & 54.0$\pm$4.2\\
    &  RCS$_\text{prob}$  &73.2$\pm$0.3	&24.2$\pm$1.3	&96.8$\pm$0.6&	88.8$\pm$1.5&	55.3$\pm$3.6\\
    
    \bottomrule
  \end{tabular}
\end{table}

\begin{table}[t]
  \centering
  \caption{Best-of-$N$ selection accuracy across settings ($N{=}20$).}
  \label{tab:ranking_n20}
  \begin{tabular}{ll|ccccc}
    \toprule
    \textbf{Model} & \textbf{Method}
    & \textbf{SciQ} & \textbf{GPQA} 
    & \textbf{Arithmetics} & \textbf{GSM8K} & \textbf{Form.Log.} 
    \\
    \midrule
    \multirow{11}{*}{Qwen2.5-3B}
    & NLL & 59.0$\pm$0.4	&16.4$\pm$2.3	&48.2$\pm$11.9	&40.6$\pm$0.5	&26.5$\pm$2.6\\
    & ANLL & 58.4$\pm$0.4	&16.2$\pm$2.6	&47.8$\pm$12.5	&40.8$\pm$1.1	&26.5$\pm$2.6\\
    & SC  & 64.9$\pm$0.6	&20.0$\pm$0.3	&95.3$\pm$5.1	&88.3$\pm$0.9	&48.1$\pm$3.6\\
    & SCW & 64.9$\pm$0.6 & 24.4$\pm$2.7 & 56.2$\pm$4.6&	33.8$\pm$3.9	&28.3$\pm$1.8\\

    & CE    & 64.2$\pm$0.9	&18.3$\pm$1.3	&95.3$\pm$5.9	&88.8$\pm$0.9	&47.1$\pm$2.4\\
    & ModeX    & 62.2$\pm$0.8	&21.9$\pm$1.7	&51.7$\pm$3.4	&37.9$\pm$2.1	&23.3$\pm$1.8\\

    & RCS$_\text{medoid}$  &65.0$\pm$0.7	&26.1$\pm$2.7	&96.5$\pm$3.5	&88.5$\pm$0.6	&47.4$\pm$4.1\\
    &RCS$_\text{uni}$& 65.5$\pm$1.0	&24.9$\pm$2.7	&96.7$\pm$2.4	&87.1$\pm$1.6	&48.1$\pm$4.4\\
    &  RCS$_\text{freq}$&64.7$\pm$0.5	&23.8$\pm$2.9&	96.3$\pm$3.3&	88.0$\pm$0.8	&47.6$\pm$5.0\\
    & RCS$_\text{cosine}$ & 65.3$\pm$0.9 & 25.2$\pm$2.9 & 94.2$\pm$1.1 & 70.4$\pm$1.8 & 42.3$\pm$3.9\\
    &  RCS$_\text{prob}$  &65.3$\pm$0.7	&24.9$\pm$2.4	&95.2$\pm$3.3	&86.6$\pm$1.1	&46.6$\pm$4.1\\
    \midrule
    
    \multirow{11}{*}{Qwen2.5-7B}
    & NLL & 66.5$\pm$0.8	&18.1$\pm$1.9	&63.3$\pm$11.2	&55.8$\pm$2.0	&30.7$\pm$4.8\\
    & ANLL & 66.5$\pm$1.0&	18.1$\pm$1.8	&63.5$\pm$12.2	&55.7$\pm$1.8	&31.2$\pm$4.4\\
    & SC  & 70.0$\pm$0.2	&26.3$\pm$0.6	&93.3$\pm$9.8	&93.1$\pm$1.3&	51.8$\pm$2.3\\
    & SCW & 70.0$\pm$0.3 & 27.6$\pm$2.0 & 56.2$\pm$1.1	&37.2$\pm$2.6	&34.9$\pm$4.0\\

    & CE    & 69.9$\pm$0.2	&25.2$\pm$0.3	&93.7$\pm$9.7	&92.6$\pm$0.8&	52.9$\pm$2.6\\
    & ModeX & 68.5$\pm$1.2 & 24.0$\pm$2.6 & 59.5$\pm$0.0 & 37.9$\pm$1.5 & 31.2$\pm$3.6\\
    & RCS$_\text{medoid}$  &69.7$\pm$0.3&	28.0$\pm$1.8	&95.0$\pm$6.9&	92.2$\pm$0.8	&52.6$\pm$1.7\\
    & RCS$_\text{uni}$&69.9$\pm$0.2	&28.7$\pm$2.2	&95.8$\pm$5.9	&91.7$\pm$1.1	&53.4$\pm$2.4\\
    &  RCS$_\text{freq}$&70.1$\pm$0.3	&28.0$\pm$1.6	&94.5$\pm$7.8	&92.7$\pm$1.1&	52.1$\pm$1.8\\
    & RCS$_\text{cosine}$ & 69.8$\pm$0.2 & 28.3$\pm$1.8 & 86.0$\pm$2.8 & 70.4$\pm$2.4 & 48.7$\pm$1.7\\
    &  RCS$_\text{prob}$  &70.0$\pm$0.3&	27.5$\pm$1.6&	95.3$\pm$6.8	&91.7$\pm$0.3	&53.4$\pm$3.2\\
    \midrule

    \multirow{11}{*}{Llama3.2-3B}
    & NLL & 53.0$\pm$0.5	&22.1$\pm$2.9	&90.2$\pm$0.8	&79.9$\pm$2.1	&39.7$\pm$0.0\\
    & ANLL & 52.8$\pm$0.9	&19.3$\pm$3.5	&91.7$\pm$0.3	&82.7$\pm$0.9	&39.9$\pm$5.3\\
    & SC  & 59.8$\pm$0.3	&15.7$\pm$0.8	&98.2$\pm$0.3	&89.3$\pm$1.3	&43.7$\pm$2.9\\
    & SCW & 60.3$\pm$0.4 & 25.0$\pm$2.6 & 67.8$\pm$1.8	&58.2$\pm$6.1	&20.9$\pm$2.0\\

    & CE    & 60.0$\pm$0.6	&17.3$\pm$1.5	&98.2$\pm$0.3	&89.7$\pm$1.3	&45.0$\pm$3.0\\
    & ModeX    & 55.0$\pm$0.7	&16.9$\pm$3.2	&73.0$\pm$0.5	&61.8$\pm$5.7	&20.1$\pm$2.0\\

    & RCS$_\text{medoid}$  &60.7$\pm$0.2&	26.1$\pm$3.4&	98.0$\pm$0.0&	89.5$\pm$1.3	&43.9$\pm$3.2\\
    & RCS$_\text{uni}$& 60.7$\pm$0.7	&26.8$\pm$2.9&	98.5$\pm$0.5	&89.5$\pm$1.6	&44.2$\pm$2.4\\
    &  RCS$_\text{freq}$&60.4$\pm$0.3	&25.7$\pm$2.9&	98.0$\pm$0.0	&89.3$\pm$1.3	&43.4$\pm$2.5\\
    & RCS$_\text{cosine}$ & 60.8$\pm$0.6 & 26.1$\pm$4.0 & 98.5$\pm$0.0 & 84.3$\pm$1.5 & 22.8$\pm$1.7\\
    &  RCS$_\text{prob}$  &60.7$\pm$0.5&	26.4$\pm$3.7	&98.0$\pm$0.0&	89.5$\pm$1.1&	42.1$\pm$2.4\\
    \midrule

    \multirow{11}{*}{Llama3.1-8B}
    & NLL & 61.0$\pm$1.5	&20.7$\pm$1.5&	81.2$\pm$4.3	&83.8$\pm$1.8&	48.9$\pm$1.8\\
    & ANLL & 61.1$\pm$1.9	&21.8$\pm$0.0	&79.8$\pm$3.9	&82.2$\pm$3.1&	44.4$\pm$2.1\\
    & SC  & 68.1$\pm$0.5	&18.9$\pm$0.4	&95.3$\pm$6.0	&91.9$\pm$1.3&	55.3$\pm$2.0\\
    & SCW & 68.3$\pm$0.1 & 32.4$\pm$0.4 & 49.5$\pm$2.3	&51.1$\pm$1.5	&27.0$\pm$1.6\\

    & CE    & 68.2$\pm$0.6	&20.7$\pm$0.0	&94.8$\pm$7.2	&91.7$\pm$1.5&	54.0$\pm$2.1\\
    & ModeX    & 62.8$\pm$0.7	&22.3$\pm$0.5	&54.8$\pm$0.6	&49.9$\pm$3.6	&24.3$\pm$3.3\\

    & RCS$_\text{medoid}$  &67.9$\pm$0.1	&35.8$\pm$0.7	&97.8$\pm$2.1	&92.2$\pm$1.8	&55.0$\pm$2.4\\
    & RCS$_\text{uni}$& 68.3$\pm$0.5	&36.0$\pm$1.1	&98.8$\pm$1.2	&91.4$\pm$1.8	&54.0$\pm$2.4\\
    &  RCS$_\text{freq}$&68.3$\pm$0.5	&33.7$\pm$0.7&	96.0$\pm$5.2&	92.2$\pm$1.8	&55.0$\pm$2.4\\
    & RCS$_\text{cosine}$ & 68.3$\pm$0.3 & 36.0$\pm$2.6 & 95.3$\pm$1.4 & 83.6$\pm$1.1 & 39.4$\pm$3.7\\
    &  RCS$_\text{prob}$  &68.1$\pm$0.3&	36.3$\pm$0.7	&97.2$\pm$1.2	&88.3$\pm$1.8	&52.9$\pm$3.0\\
    \midrule

    \multirow{11}{*}{Gemma2-9B}
    & NLL & 74.1$\pm$0.3	&27.1$\pm$1.1&	96.2$\pm$0.3&	89.3$\pm$0.9&	52.1$\pm$0.9\\
    & ANLL & 64.8$\pm$0.5&	23.0$\pm$1.1	&96.3$\pm$0.3	&89.8$\pm$0.9	&54.5$\pm$1.7\\
    & SC  & 74.1$\pm$0.8	&22.1$\pm$2.0&	97.2$\pm$0.3&	91.7$\pm$0.3	&54.0$\pm$0.8\\
    & SCW & 74.1$\pm$0.5 & 28.8$\pm$2.0 & 93.2$\pm$0.6	&63.6$\pm$2.0	&50.5$\pm$1.7\\

    & CE    & 74.3$\pm$0.4&	21.8$\pm$0.5&	97.2$\pm$0.3&	91.7$\pm$0.8&	54.0$\pm$0.8\\
    & ModeX    & 73.0$\pm$0.4&22.1$\pm$1.6	&92.3$\pm$2.0&	64.8$\pm$1.1	&51.9$\pm$5.4\\
    & RCS$_\text{medoid}$  &74.2$\pm$0.4	&28.7$\pm$1.7&	97.2$\pm$0.3&	91.7$\pm$0.3&	54.2$\pm$0.5\\
    & RCS$_\text{uni}$& 74.7$\pm$0.4	&29.9$\pm$1.3	&97.2$\pm$0.3	&91.7$\pm$0.3	&54.2$\pm$0.5\\
    &  RCS$_\text{freq}$&74.2$\pm$0.6&	29.4$\pm$2.1&	97.2$\pm$0.3	&91.9$\pm$0.0&	54.2$\pm$0.5\\
    & RCS$_\text{cosine}$ & 74.6$\pm$0.4 & 29.0$\pm$0.5 & 97.2$\pm$0.3 & 76.3$\pm$0.8 & 52.1$\pm$0.9\\
    &  RCS$_\text{prob}$  &74.6$\pm$0.6	&29.5$\pm$0.9	&97.2$\pm$0.3	&91.5$\pm$0.3	&54.2$\pm$0.9\\
    
    \bottomrule
  \end{tabular}
\end{table}

\begin{table}[t]
  \centering
  \caption{Best-of-$N$ selection accuracy across settings ($N{=}40$).}
  \label{tab:ranking_n40}
  \begin{tabular}{ll|ccccc}
    \toprule
    \textbf{Model} & \textbf{Method}
    & \textbf{SciQ} & \textbf{GPQA} 
    & \textbf{Arithmetics} & \textbf{GSM8K} & \textbf{Form.Log.} 
    \\
    \midrule
    \multirow{11}{*}{Qwen2.5-3B}
    & NLL & 56.3$\pm$0.8	&16.8$\pm$1.8	&45.3$\pm$13.0&	42.8$\pm$1.5&	25.9$\pm$3.8\\
    & ANLL & 56.1$\pm$0.9	&16.8$\pm$2.0	&44.8$\pm$13.3	&43.0$\pm$2.8&	25.1$\pm$3.8\\
    & SC  & 	65.2$\pm$0.4&	21.8$\pm$1.4	&97.3$\pm$2.9&	90.2$\pm$1.1&	49.5$\pm$1.2\\
    & SCW & 65.3$\pm$0.3 & 25.6$\pm$2.3 & 54.5$\pm$4.2	&38.1$\pm$1.8&	30.7$\pm$3.0\\

    & CE    & 65.2$\pm$0.3	&20.4$\pm$2.7	&97.5$\pm$2.6	&90.4$\pm$0.5	&48.9$\pm$3.2\\
    & ModeX    & 63.0$\pm$1.0	&20.0$\pm$0.5&	49.8$\pm$3.2	&35.7$\pm$3.1&	25.7$\pm$6.7\\
    & RCS$_\text{medoid}$  &65.5$\pm$0.5&	25.0$\pm$1.3	&98.2$\pm$1.5	&90.0$\pm$0.8	&47.9$\pm$1.2\\
    & RCS$_\text{uni}$& 65.7$\pm$0.7&	25.0$\pm$1.3	&98.5$\pm$1.3	&89.7$\pm$1.3	&48.4$\pm$0.8\\
    &  RCS$_\text{freq}$&65.3$\pm$0.5	&23.8$\pm$2.4&	97.7$\pm$2.4&	90.0$\pm$0.8	&47.6$\pm$2.1\\
    & RCS$_\text{cosine}$ & 65.7$\pm$0.7 & 25.7$\pm$0.6 & 97.2$\pm$1.1 & 76.1$\pm$2.2 & 44.7$\pm$1.2\\
    &  RCS$_\text{prob}$  &65.9$\pm$0.7&	24.4$\pm$1.0	&98.2$\pm$1.6&	89.3$\pm$1.3	&47.4$\pm$0.5\\
    \midrule
    
    \multirow{11}{*}{Qwen2.5-7B}
    & NLL & 66.0$\pm$0.4	&18.5$\pm$1.2&	63.0$\pm$10.1	&57.5$\pm$3.1&	30.7$\pm$4.4\\
    & ANLL & 65.9$\pm$0.6&	18.5$\pm$0.8	&62.8$\pm$9.3	&57.5$\pm$3.4&	30.4$\pm$3.2\\
    & SC  & 70.5$\pm$0.3	&26.1$\pm$1.6&	96.2$\pm$5.8	&94.4$\pm$0.5&	54.0$\pm$0.8\\
    & SCW & 70.4$\pm$0.1 & 28.2$\pm$0.6 & 50.2$\pm$1.8&	39.1$\pm$4.8	&32.8$\pm$2.3\\

    & CE    & 70.4$\pm$0.6&	26.9$\pm$3.1	&96.3$\pm$5.5	&94.4$\pm$0.0&	53.7$\pm$1.2\\
    & ModeX & 69.6$\pm$0.3 & 25.0$\pm$1.7 & 53.2$\pm$1.8 & 38.6$\pm$1.8 & 31.2$\pm$2.6\\
    & RCS$_\text{medoid}$  &70.0$\pm$0.1	&28.0$\pm$0.9	&97.2$\pm$4.0&	94.8$\pm$0.3&	53.2$\pm$2.1\\
    & RCS$_\text{uni}$&70.3$\pm$0.4	&29.0$\pm$0.9&	98.2$\pm$2.3&	94.6$\pm$0.3	&52.6$\pm$3.7\\
    &  RCS$_\text{freq}$&70.5$\pm$0.2	&27.5$\pm$0.5&	96.5$\pm$5.2	&94.6$\pm$0.3&	52.9$\pm$3.2\\
    & RCS$_\text{cosine}$ & 70.2$\pm$0.4 & 28.3$\pm$0.3 & 94.0$\pm$0.0 & 75.6$\pm$1.8 & 50.3$\pm$1.7\\
    &  RCS$_\text{prob}$  &70.4$\pm$0.4	&27.3$\pm$0.8&	97.5$\pm$3.5	&94.6$\pm$0.3&	52.4$\pm$3.5\\
    \midrule

    \multirow{11}{*}{Llama3.2-3B}
    & NLL & 53.4$\pm$0.3&	21.6$\pm$1.7	&93.5$\pm$2.6&	82.7$\pm$1.0&	36.2$\pm$4.0\\
    & ANLL & 52.8$\pm$0.6	&18.7$\pm$2.3&	93.3$\pm$2.5&	83.2$\pm$1.8&	41.8$\pm$5.7\\
    & SC  & 60.4$\pm$0.7	&17.6$\pm$0.9&	98.7$\pm$0.6&	91.2$\pm$0.8&	44.2$\pm$0.5\\
    & SCW & 60.6$\pm$0.9 & 24.7$\pm$2.4 & 69.0$\pm$4.2&	58.0$\pm$2.6	&19.6$\pm$3.0\\

    & CE    & 60.2$\pm$0.4&	21.2$\pm$3.2	&98.8$\pm$0.3&	91.2$\pm$0.6&	43.1$\pm$0.9\\
    & ModeX    & 55.8$\pm$0.9	&19.0$\pm$2.9&	72.5$\pm$2.6	&57.0$\pm$4.1	&22.0$\pm$3.2\\
    & RCS$_\text{medoid}$  &60.9$\pm$1.0	&29.4$\pm$1.6&	98.7$\pm$0.6&	91.2$\pm$0.6&	45.2$\pm$0.8\\
    & RCS$_\text{uni}$&	60.9$\pm$0.8	&29.4$\pm$2.0	&98.5$\pm$0.9	&90.4$\pm$0.5&	46.3$\pm$1.7\\
    &  RCS$_\text{freq}$&61.0$\pm$1.0&	26.9$\pm$1.9&	98.7$\pm$0.6&	90.9$\pm$0.5	&44.2$\pm$1.7\\
    & RCS$_\text{cosine}$ & 61.1$\pm$1.1 & 29.9$\pm$2.0 & 98.8$\pm$1.1 & 86.3$\pm$0.5 & 23.0$\pm$2.9\\
    &  RCS$_\text{prob}$  &61.0$\pm$0.6	&29.4$\pm$2.1	&98.3$\pm$0.3&	90.7$\pm$0.3&	44.7$\pm$2.8\\
    \midrule

    \multirow{11}{*}{Llama3.1-8B}
    & NLL & 61.8$\pm$0.9	&18.9$\pm$1.1&	83.7$\pm$1.2	&83.9$\pm$2.5&	46.8$\pm$2.1\\
    & ANLL & 61.3$\pm$0.8	&20.5$\pm$1.8&	84.0$\pm$1.7&	81.7$\pm$2.0&	43.4$\pm$4.0\\
    & SC  & 69.3$\pm$0.3	&19.7$\pm$0.7&	96.3$\pm$4.2&	93.2$\pm$0.8&	51.9$\pm$2.0\\
    & SCW & 69.2$\pm$0.4 & 25.6$\pm$1.1 & 54.2$\pm$1.0&	52.6$\pm$1.2	&27.0$\pm$2.9\\

    & CE    & 68.5$\pm$0.4&	18.9$\pm$0.4&	96.5$\pm$4.3&	93.2$\pm$0.8&	51.1$\pm$1.7\\
    & ModeX    & 64.8$\pm$0.3&	19.3$\pm$2.7&	53.5$\pm$0.9&	51.4$\pm$1.6	&23.0$\pm$2.1\\
    & RCS$_\text{medoid}$  &69.1$\pm$0.3	&31.6$\pm$0.7&	98.5$\pm$0.5&	92.4$\pm$0.5	&51.1$\pm$1.2\\
    & RCS$_\text{uni}$& 69.6$\pm$0.2	&31.4$\pm$0.4	&98.7$\pm$1.0&	92.2$\pm$0.3&	50.5$\pm$1.2\\
    &  RCS$_\text{freq}$&69.2$\pm$0.8&	28.2$\pm$1.8	&96.7$\pm$3.6&	93.1$\pm$0.6&	51.1$\pm$1.2\\
    & RCS$_\text{cosine}$ & 69.5$\pm$0.2 & 31.9$\pm$1.1 & 99.2$\pm$0.6 & 84.9$\pm$0.8 & 38.1$\pm$1.4\\
    &  RCS$_\text{prob}$  &69.6$\pm$0.4&	29.5$\pm$1.5&	98.5$\pm$0.9&91.9$\pm$1.3	&51.6$\pm$2.1\\
    \midrule

    \multirow{11}{*}{Gemma2-9B}
    & NLL & 74.2$\pm$0.1	&26.8$\pm$2.1&	96.0$\pm$0.5	&89.7$\pm$0.8&51.3$\pm$3.9\\
    & ANLL & 62.2$\pm$1.5	&20.7$\pm$1.6&	96.2$\pm$0.3&	89.5$\pm$1.8&	53.7$\pm$2.0\\
    & SC  & 	74.2$\pm$0.4	&22.5$\pm$0.8	&97.5$\pm$0.0	&91.7$\pm$0.8&	55.6$\pm$0.8\\
    & SCW & 74.2$\pm$0.5 & 26.9$\pm$0.9 & 90.3$\pm$2.1	&64.8$\pm$1.1&	50.5$\pm$3.3\\

    & CE    & 74.2$\pm$0.3	&19.3$\pm$0.6&	97.5$\pm$0.0	&91.5$\pm$0.3	&55.3$\pm$1.2\\
    & ModeX    & 71.9$\pm$0.4 &	21.4$\pm$3.2&	93.5$\pm$0.5	&66.3$\pm$2.9	&54.0$\pm$2.4\\
    & RCS$_\text{medoid}$  &74.4$\pm$0.4&	29.4$\pm$1.2	&97.5$\pm$0.0	&91.7$\pm$0.8&	55.0$\pm$0.9\\
    & RCS$_\text{uni}$& 74.4$\pm$0.8	&29.7$\pm$1.6&	97.5$\pm$0.0&	91.5$\pm$0.6	&54.8$\pm$0.8\\
    &  RCS$_\text{freq}$&74.2$\pm$0.4&	28.0$\pm$1.0	&97.5$\pm$0.0	&91.7$\pm$0.8	&55.3$\pm$1.2\\
    & RCS$_\text{cosine}$ & 74.3$\pm$0.6 & 29.9$\pm$0.8 & 97.5$\pm$0.0 & 76.7$\pm$0.9 & 52.6$\pm$0.5\\
    &  RCS$_\text{prob}$  &74.4$\pm$0.5&	29.7$\pm$0.8	&97.3$\pm$0.3&	91.2$\pm$0.8&	55.3$\pm$0.5\\
    
    \bottomrule
  \end{tabular}
\end{table}


\end{document}